\documentclass[journal]{IEEEtran}

\usepackage{multirow}
\usepackage{bm}
\usepackage{color}
\usepackage{epsfig}
\usepackage{graphicx}
\usepackage{amsmath}
\usepackage{amssymb}
\usepackage{xspace}
\usepackage{algorithm} %format of the algorithm
\usepackage{algorithmic} %format of the algorithm
\usepackage[dvipsnames, svgnames, x11names, table]{xcolor}

\usepackage{amsthm}
\usepackage{array}
\usepackage{colortbl}
\usepackage{bbding}
\usepackage{url}
\usepackage{hyperref}
\usepackage{booktabs}
\usepackage{multirow}
\usepackage{cite}
\usepackage{graphicx}
\usepackage{tabularx}
\usepackage{epstopdf}
\usepackage{subfigure}
\usepackage[normalem]{ulem}
\useunder{\uline}{\ul}{}

\theoremstyle{remark}

\newcommand{\eg}{\textit{e.g.,~}}
\newcommand{\ie}{\textit{i.e.,~}}
\newcommand{\cf}{\textit{c.f.~}}

\newcommand{\etal}{\textit{et al.~}}

\begin{document}

\title{Generalized Manifold Mixup for Efficient Representation Learning}
\title{Dynamic Feature Augmentation via Generalized Manifold Mixup}
% \title{Generalized Manifold Mixup}
\title{Dynamic Feature Augmentation for Visual Categorization}                     
\title{ShuffleMix: Regularizing Visual Representation Learning with Soft Dropout of Hidden States} 
\title{ShuffleMix: Improving Representations via Channel-Wise Shuffle of Interpolated Hidden States} 

\author{Kangjun Liu, Ke Chen,~\IEEEmembership{Member,~IEEE,} Lihua Guo, Yaowei Wang,~and~Kui Jia,~\IEEEmembership{Member,~IEEE}
%\thanks{$^*$equal contribution to this work; $^\dagger$corresponding author.}
\IEEEcompsocitemizethanks{
\IEEEcompsocthanksitem This work is supported in part by the National Natural Science Foundation of China (Grant No.: 61771201, 61902131), the Program for Guangdong Introducing Innovative and Enterpreneurial Teams (Grant No.: 2017ZT07X183), the Guangdong Provincial Key Laboratory of  Human Digital Twin (Grant No.: 2022B1212010004), the Guangdong Basic and Applied Basic Research Foundation (Grant No.: 2022A1515011549). %K. Chen and K. Jia are the corresponding authors of this work.
\IEEEcompsocthanksitem K. Liu is with the Shien-Ming Wu School of Intelligent Engineering, South China University of Technology, Guangzhou 510641, China. K. Chen, L. Guo and K. Jia are with the School of Electronic and Information Engineering, South China University of Technology, Guangzhou 510641, China.
Y. Wang is with the Peng Cheng Laboratory, Shenzhen, China. }
% E-mails: chenk@scut.edu.cn; kuijia@scut.edu.cn.}% <-this % stops a space}
}
	
\markboth{Journal of \LaTeX\ Class Files,~Vol.~X, No.~X, May~2022}%
{Shell \MakeLowercase{\textit{et al.}}: Bare Demo of IEEEtran.cls for IEEE Journals}	

\maketitle

% \IEEEcompsoctitleabstractindextext{

\begin{abstract}

Mixup style data augmentation algorithms have been widely adopted in various tasks as implicit network regularization on representation learning to improve model generalization, which can be achieved by a linear interpolation of labeled samples in input or feature space as well as target space.
Inspired by good robustness of alternative dropout strategies against over-fitting on limited patterns of training samples, this paper introduces a novel concept of ShuffleMix -- {\em Shuffle of Mixed hidden features}, which can be interpreted as a kind of dropout operation in feature space.
Specifically, our ShuffleMix method favors a simple linear shuffle of randomly selected feature channels for feature mixup in-between training samples to leverage semantic interpolated supervision signals, which can be extended to a generalized shuffle operation via additionally combining linear interpolations of intra-channel features. 
Compared to its direct competitor of feature augmentation -- the Manifold Mixup, the proposed ShuffleMix can gain superior generalization, owing to imposing more flexible and smooth constraints on generating samples and achieving regularization effects of channel-wise feature dropout. 
Experimental results on several public benchmarking datasets of single-label and multi-label visual classification tasks can confirm the effectiveness of our method on consistently improving representations over the state-of-the-art mixup augmentation.

\end{abstract}

\begin{IEEEkeywords}
Data augmentation, Image classification, Representation learning, Multi-label classification, Feature shuffle.
\end{IEEEkeywords}

\IEEEpeerreviewmaketitle

%-----------------------------------------------------------------------------------------
\section{Introduction}\label{sec:intro}

\IEEEPARstart{D}{ata} augmentation is widely used in the problem of visual recognition \cite{deng2009imagenet, Krizhevsky2012ImageNetCW,Szegedy2016RethinkingTI,he2016deep, Zhang2018MultilabelIC, Chen2019MultiLabelIR}, which allows training deep neural models with not only the original data but also extra data after proper transformation operations (\eg cropping, jittering). 
Performance gain using data augmentation on deep neural networks can be explained by enriching data diversity in original samples' neighborhood to impose low-dimensional manifolds in high-dimensional representation space \cite{lei2020geometric,lei2019geometric}, which can thus be viewed as a class of implicit regularization on feature encoding \cite{Dai2021ImplicitDA}.
As a result, enriching data with different kinds of augmentation strategies can effectively improve model generalization and prevent training of image classifiers from over-fitting on limited patterns of training samples.
Recently, the pioneering Mixup method \cite{zhang2018mixup} and its follow-upper \cite{Guo2019MixUpAL} have attracted wide attention via augmenting data by linear interpolations of a random pair of examples' raw input and semantically interpolated labels for training deep models.%, which can obtain smoother decision boundaries of image classification.

% \begin{figure}[t]
%     \centering
%     \includegraphics[width=\linewidth]{figures/smooth_boundary_3classes_2.pdf}
%     \caption{Decision boundaries for different augmentation methods on a synthetic dataset for 3-class classification, which are generated based on the middle circle class. 
%     The darker the color, the higher the probability of that region belonging to the middle class. 
%     We can observe that our ShuffleMix can have smoother decision boundaries and more precise representation manifolds than those of the Input Mixup \cite{zhang2018mixup} and the Manifold Mixup \cite{Verma2019ManifoldMB}.
%     Classification accuracies of three methods are shown in the bottom right of each plot. 
%     More details are presented in Sec. \ref{sec:visualization}. (Best viewed in colors)
%     \textcolor{red}{Ke: merge fig.1 and 2 into one figure.}
%     }
%         \label{fig:smooth_boundary_3classes}
% % \vspace{-0.2cm}
% \end{figure}

\begin{figure}[t]
    \centering
    \subfigure[Input Mixup]{ 
            \includegraphics[width=0.365\linewidth]{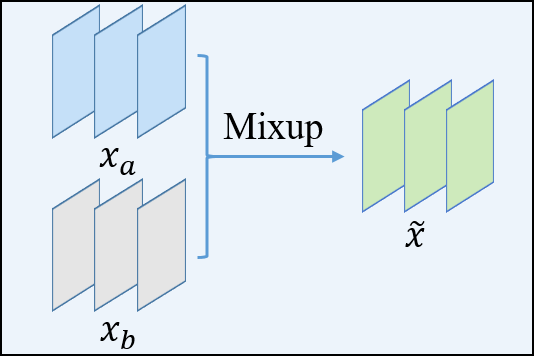}
            \label{fig:pipeline_0}}
     \quad      
     \hspace{2mm} 
     \subfigure[Manifold Mixup]{ 
             \includegraphics[width=0.48\linewidth]{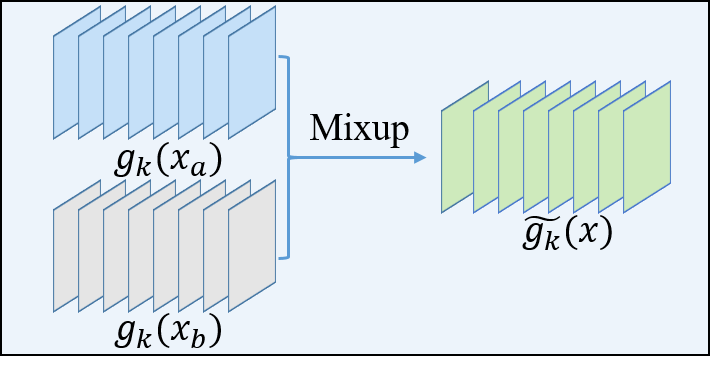}
             \label{fig:pipeline_1}}
     \quad
%    \subfigure[Hard ShuffleMix]{
%            \includegraphics[width=0.86\linewidth]{figures/pipeline_2_1.png}
%            \label{fig:pipeline_2}}
%    \quad 
    \subfigure[ShuffleMix]{
            \includegraphics[width=0.98\linewidth]{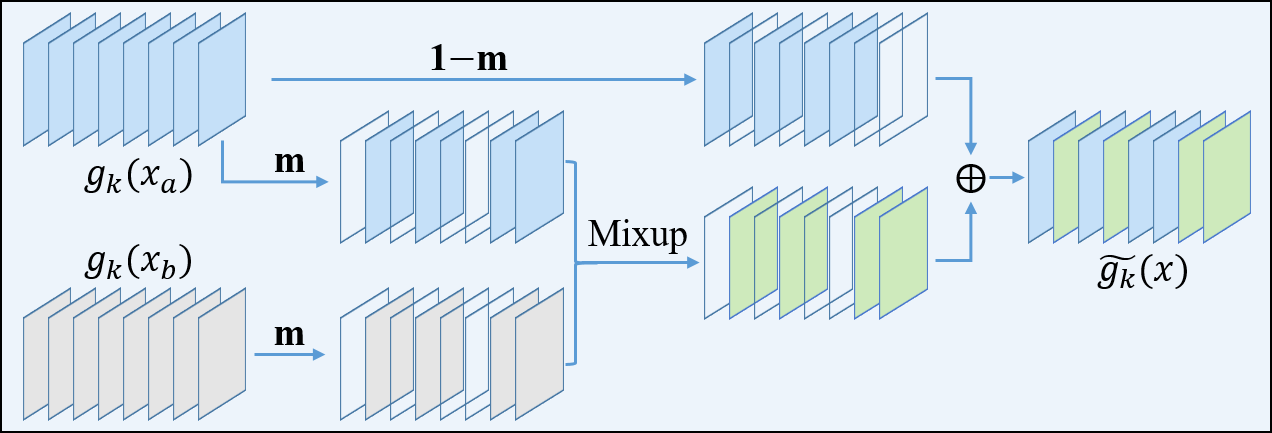}
            \label{fig:pipeline_3}}
	\quad           
    \subfigure[Decision Boundaries]{	
    		\includegraphics[width=\linewidth, trim=13 10 12 10, clip]{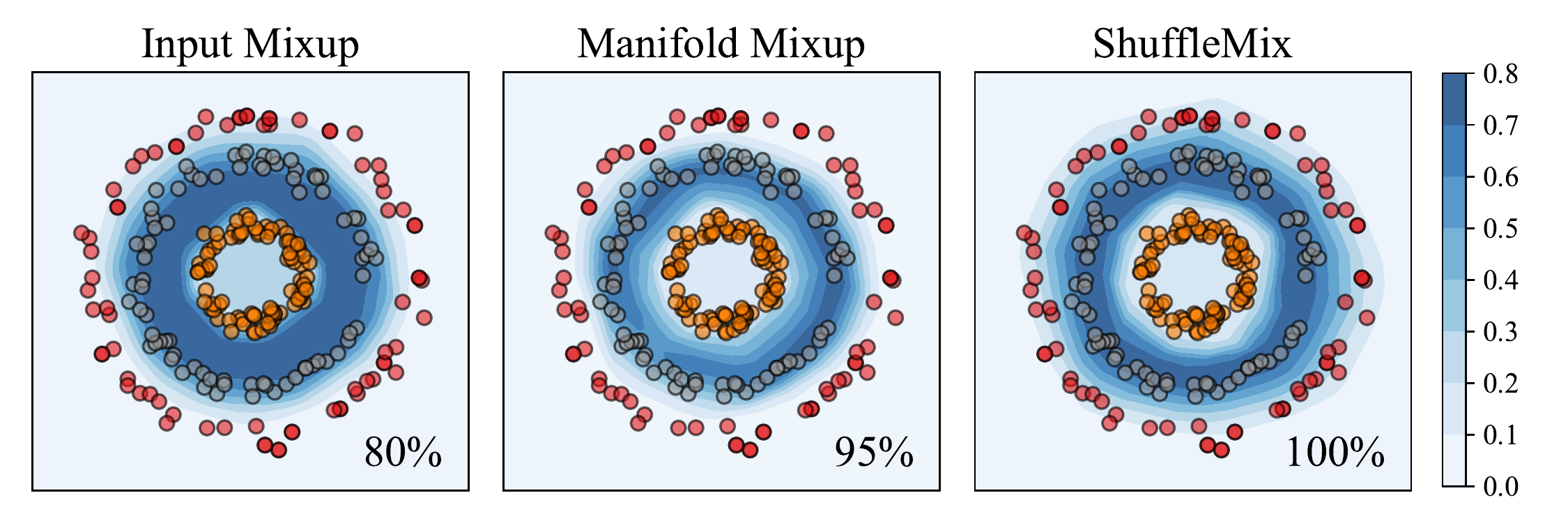} 
            \label{fig:boundaries}}
	\quad 
	\vspace{-2mm} 
    \caption{Comparison with the Input Mixup \cite{zhang2018mixup}, the Manifold Mixup \cite{Verma2019ManifoldMB} and our {\em Shuffle Mixup of Neural Features from Hidden Layers} (ShuffleMix), where $x$ denotes the input image, $\Tilde{x}$ is the mixed input image, ${g_k}(x)$ denotes the hidden representations of neural networks and $\Tilde{g_k}(x)$ is the combined representation. 
    Our ShuffleMix in (c) are more flexible than Manifold Mixup in (b), which can be viewed as a special case of our ShuffleMix.
    While, decision boundaries in (d) are generated based on the middle circle class for different augmentation methods on a synthetic dataset of 3-class classification. 
    The darker the color, the higher the probability of that region belonging to the middle class. 
    We can observe that our ShuffleMix can have smoother decision boundaries and better representation manifolds (\ie ratio of samples falling into the darkest region) than those of the Input Mixup \cite{zhang2018mixup} and the Manifold Mixup \cite{Verma2019ManifoldMB}.
    Classification accuracies of three methods are shown in the bottom right of each plot. (Best viewed in colors) 
    }
    \label{fig:pipline}\vspace{-0.3cm}
\end{figure}

% SpatialDropout \cite{tompson2015efficient}
Alternatively, random dropout strategies on hidden states \cite{Srivastava2014DropoutAS} or spatial regions \cite{tompson2015efficient, Ghiasi2018DropBlockAR} have been investigated to enforce neural models to distribute the knowledge learned from data to a large size of inter-neuron connection, rather than focusing on a limited portion. 
A number of recent works \cite{wei2020implicit, mianjy2018implicit, Liang2021RDropRD} have shown that introduction of random feature removal operations into representation learning as a kind of implicit network regularization can improve model generalization and representation quality. 
Moreover, dropout as a noise scheme can be also regarded as a form of implicit data augmentation  \cite{Konda2015DropoutAD}, which theoretically maps the sub-regions of both input space and all high-probability natural distribution to label space.
% In the context of mixup-style data augmentation, a number of algorithms based on regional dropout have been proposed, \eg Cutout \cite{Devries2017ImprovedRO} and CutMix \cite{yun2019cutmix}, which are able to enhance the model generalization owing to discovering discriminative localized features from less informative parts of object. 
In this regard, many mixup-style data augmentation algorithms based on regional dropout have been proposed, \eg Cutout \cite{Devries2017ImprovedRO}, CutMix \cite{yun2019cutmix} and Random Erasing \cite{zhong2020random}, which are able to enhance model generalization owing to discovering discriminative local features from less informative parts of object. 
However, these existing augmentation methods are just designed to impose the spatial attention mechanism into representation learning via regional dropout-like operations in the input space, without exploiting latent structure of representation manifolds.

The first exploration of mixup-style augmentation to feature space is the Manifold Mixup \cite{Verma2019ManifoldMB}, which is designed on linear interpolations of hidden states of neural networks at multiple levels
%In this work, we are interested in the Manifold Mixup \cite{Verma2019ManifoldMB} since they have demonstrated 
%Such a design can encourage better representations with 
for smooth boundaries and flatter representations during model training.
%avoid label ambiguities of augmented samples.  
%\textcolor{black}{However, direct linear combination of different sample representations in hidden states could lead to a misunderstanding of the third part category, which could not be simply represented by a linear mixed labels (cf. Figure \ref{fig:smooth_boundary_3classes} for an illustration).}
Inspired by the ShuffleNet \cite{Zhang2018ShuffleNetAE}, we propose a simple yet effective alternative of mixup augmentation in feature space -- ShuffleMix, which favors for a channel-wise shuffle operation on feature maps from hidden layers of neural networks.
%, and can empirically alleviate the problems mentioned above.
Technically, a randomly-generated binary index vector $\bm{m}$ is employed for selecting channel dimensions to be shuffled in-between feature maps of two samples, which can intuitively be interpreted as a linear interpolation of feature maps after channel-wise dropout. 
Beyond leveraging semantically interpolated supervision signals, such a hard shuffle augmentation of hidden states can improve generalization of visual representations owing to its implicit nature of ensembling models sharing parameters as the dropout \cite{Srivastava2014DropoutAS}.
% Our hard shuffle mixup can be further extended to a soft shuffle mixup via replacing a part of randomly-selected feature channels with mixup interpolated features with another data point as the Manifold Mixup \cite{Verma2019ManifoldMB}, which can be considered as soft dropout via a generalized linear combination of feature maps.
Our hard shuffle mixup can be further extended to a soft shuffle mixup via replacing a part of randomly-selected feature channels with linear interpolated features, similar to the Manifold Mixup \cite{Verma2019ManifoldMB}.
% Compared with dropout \cite{Srivastava2014DropoutAS}, this shuffle mixup operation can be considered as soft dropout via a generalized linear combination of feature maps and noises.
As a result, our soft ShuffleMix can not only inherit the advantages of randomly dropout on generalization from the hard ShuffleMix, but also exploit representation manifolds in the more flexible way, as the Manifold Mixup can be considered as its special case (\cf Section \ref{subsec:shufflemix} for details).
%Owing to mixing randomly-selected feature channels, our soft ShuffleMix can inherit mixed features from falling into the third category by retaining enough original feature maps.
% Compared with the vanilla Dropout \cite{Srivastava2014DropoutAS}, this shuffle mixup operation can be considered as soft dropout via a generalized linear combination of feature maps.
%Therefore, the proposed ShuffleMix can be more flexible than existing Manifold Mixup for feature augmentation, as it can treat the Manifold Mixup as its special case.
Comparison of the Input Mixup \cite{zhang2018mixup}, the Manifold Mixup \cite{Verma2019ManifoldMB} and the proposed ShuffleMix are illustrated in \textcolor{black}{Fig. \ref{fig:pipline}}.
%Meanwhile, the proposed ShuffleMix operation can not only explicitly exploit latent label correlation in the feature space via a generalized linear combination of feature maps, but also be considered as soft dropout with interpolated schemes for preventing over-fitting.

Recently, superior robustness can be gained via incorporating noise injection schemes into feature interpolations of data points as the Noisy Feature Mixup (NFM) \cite{Lim2021NoisyFM}, which are tolerant to data perturbations in their neighboring feature space.
%Encouraged by such a success, 
% Moreover, similar to the noise injection schemes introduced in \cite{Lim2021NoisyFM,Franceschi2018RobustnessOC}, 
% our ShuffleMix method can further improve representation robustness via explicitly exploring data perturbations with noise injection schemes in the neighbourhood of feature space.
Similarly, our ShuffleMix method can further improve representation robustness via explicitly exploring data perturbations with noise injection both in the neighbourhood of original feature maps and interpolated schemes.
%Consequently, our ShuffleMix can not only impose smoothness regularization into representation learning in an implicit manner as existing feature augmentation methods \cite{Verma2019ManifoldMB, Lim2021NoisyFM, Franceschi2018RobustnessOC}, but also alleviate the risk of over-fitting owing to our design of dropout-style feature augmentation. 
Experiment results on popular benchmarks of single-label and multi-label image classification tasks verify that the proposed ShuffleMix can consistently outperform existing mixup-style data augmentation methods including the state-of-the-art NFM \cite{Lim2021NoisyFM}.

Main contributions of this paper are as follows.
\begin{itemize}
\item This paper proposes a novel ShuffleMix feature augmentation method, which is simple yet generic to existing image classification networks.
\item Technically, the proposed ShuffleMix is designed via a generalized linear combination of channel-wise features of a pair of randomly selected data samples, which can improve model generalization and robustness owing to its soft dropout characteristics and more flexibility to exploit representation manifolds. 
%\item We successfully apply the proposed method to several different visual classification tasks, \ie robustness to input noise injection, fine-grained classification and multi-label classification, with some appropriate modifications.
\item We extensively conduct experiments on several popular benchmarks of different visual classification tasks, whose results can demonstrate the effectiveness of the proposed ShuffleMix and superior performance to the state-of-the-art mixup-style augmentation methods.
\end{itemize}
Source codes and pre-trained models will be released after acceptance\footnote{\url{Link-To-Download-Source-Codes-and-Pre-Trained-Models.}}.

\section{Related work}\label{sec:relatedWorks}

% \textcolor{black}{Ke: I do not have time to check the text here.}
In this section, we firstly briefly review some recent works regrading mixup style data augmentation. Secondly, existing works about implicit feature regularization in model training are investigated. Finally, we provide a brief survey about representative works in single-label and multi-label image recognition. 

\vspace{0.1cm}\noindent\textbf{Mixup Style Data Augmentation --}
With the Mixup \cite{zhang2018mixup} verifying the effectiveness of data augmentation by interpolating two input images, there appears an amount of mixup style data augmentation methods \cite{Tokozume2018BetweenClassLF,Guo2019MixUpAL,Verma2019ManifoldMB, yun2019cutmix,Lee2020SmoothMixAS,kim2020puzzle,uddin2020saliencymix,Huang2021SnapMixSP}.
Guo \etal developed a theoretical understanding for Mixup as a kind of out-of-manifold regularization and proposed an adaptive Mixup (AdaMixup) with learnable mixing polices.
Meanwhile, Yun \etal \cite{yun2019cutmix} proposed a CutMix augmentation strategy to blend two input images with patch-based spatial region shuffle instead of a linear interpolation. 
Based on the CutMix, Kim \etal \cite{kim2020puzzle} proposed a PuzzleMix, and Uddin \etal \cite{uddin2020saliencymix} proposed a SaliencyMix, both of which aimed to detect the representative regions for mixing with the help of saliency maps.
Lee \etal \cite{Lee2020SmoothMixAS} proposed a SmoothMix augmentation by performing the blending of two input images with a randomly generated mask for weighting instead of hard shuffle, while Huang \etal \cite{Huang2021SnapMixSP} proposed a SnapMix via obtaining semantic-relatedness proportion for mixed regions with the help of saliency maps.
PuzzleMix \cite{kim2020puzzle},  SaliencyMix \cite{uddin2020saliencymix}, SmoothMix \cite{Lee2020SmoothMixAS} and SnapMix \cite{Huang2021SnapMixSP} all are the variants of CutMix \cite{yun2019cutmix} with different region shuffle strategies and interpolated weights of labels. 
% More detailed reviews about mixup style data augmentation are available at \cite{Lewy2021AnOO}.
Different from above works, Verma \etal \cite{Verma2019ManifoldMB} proposed a Manifold Mixup augmentation strategy by extending the blending of input images to the interpolation of arbitrary hidden feature representations for imposing smooth decision boundaries, which is the direct competitor with our ShuffleMix.
\textcolor{black}{Recently, several more feature augmentation methods are proposed to improve representation learning, \eg AlignMix \cite{Venkataramanan2021AlignMixIR} with alignment of feature tensors and Shuffle Augmentation of Features (SAF) \cite{Xu2022ShuffleAO} with augmenting source features from target distribution.}
Different from existing feature augmentation algorithms, our ShuffleMix augmentation, which can be interpreted as a kind of dropout operation, can benefit from spreading discrimination of representations to more feature dimensions, rather than overfitting on a limited size of patterns.

%Owing to a more flexible combination of hidden representations, our ShuffleMix can further improve model generalization over existing .
% \textcolor{black}{Ke: add the statement of alignMix and shuffle augmentation of features here.}

\vspace{0.1cm}\noindent\textbf{Implicit Feature Regularization --}
Regularization is generally used as an important technique to prevent over-fitting in deep learning. 
Except for the classical regularization methods, \eg the Weight Decay \cite{Loshchilov2019DecoupledWD}, the Dropout \cite{Srivastava2014DropoutAS} and the Batch Normalization \cite{Ioffe2015BatchNA}, Hernandez-Garcia \etal \cite{HernndezGarca2018DataAI} demonstrated that data augmentation can be treated as an implicit regularization, which can perform better than those classical regularization for model training.
\textcolor{black}{
More works explored different type of dropout for regularizing convolutional networks, including the DropBlock \cite{Ghiasi2018DropBlockAR} and the Weighted Channel Dropout \cite{Hou2019WeightedCD}.
Specially, traditional dropout style methods are all regularizing the input images and features by replacing some spatial regions or dimensions with zero, which suffers from  losing important information.
}
Carratino \etal \cite{Carratino2020OnMR} and Zhang \etal \cite{zhang2020does} further analyzed that Mixup \cite{zhang2018mixup} can help to improve the performance of models as a specific regularization based on the Taylor approximation. 
Inspired by such an observation, the proposed ShuffleMix can be considered as complementing mixup augmentation with a dropout like operation as implicit feature regularization in a unified framework.

%Robustness as another important indicator of model, refers to the fact that models need to be able to identify adversarial samples \cite{kurakin2016adversarial} or noisy samples \cite{Fawzi2016RobustnessOC}. 
%Beckham \etal \cite{Beckham2019OnAM} proposed a strategy of blending adversarial sample and authentic sample for improving the robustness over adversarial samples.
%Lim \etal \cite{Lim2021NoisyFM} attempt to improve model robustness to input noises with the combination of Manifold Mixup \cite{Verma2019ManifoldMB} and noise injection, which encourages us to combine our ShuffleMix and noise injection for better robustness.

\vspace{0.1cm}\noindent\textbf{Single-Label Visual Classification --}
% \textcolor{red}{add image classification text here, you can organize the text here without dividing into general and fine-grained image classification, but in a mixed manner.}
The problem of single-label image recognition has been actively studied for decades. 
Existing works are generally categorized into part-based methods \cite{huang2016part, fu2017look, zheng2019learning} and part-free methods \cite{Krizhevsky2012ImageNetCW,Szegedy2016RethinkingTI,he2016deep, du2020fine, Chang2020TheDI, Shi2019FineGrainedIC, Ding2021APCNNWS}.
Part-based methods usually attempt to learn critical information from detected key parts region of objects for fine-grained recognition. 
Huang \etal \cite{huang2016part} proposed a Part-Stacked CNN for learning distinguished information via modeling the subtle differences between objects parts. 
To better locate the object parts, Zheng \etal \cite{zheng2019learning} proposed a progressive-attention CNN to locate object parts with varying scale. 
Unlike part-based methods, part-free methods always focus on extracting discriminative features from global region via exploiting training strategies or network constraints.
Du \etal \cite{du2020fine} proposed a progressive training strategy and a jigsaw puzzle operation to exploit multi-granularity features learning.
Shi \etal \cite{Shi2019FineGrainedIC} attempted to exploit hierarchical label structure of fine-grained classes by leveraging a cascaded softmax loss and a generalized large-margin loss.
Ding \etal \cite{Ding2021APCNNWS} proposed an attention pyramid CNN to enhance feature representation by integrating cross-level feature information.
Our ShuffleMix aims to enhance model generalization via implicit regularization on representation learning, with combining both mixup style augmentation and dropout operations in a unified framework, which can thus be readily adopted in existing classification networks. 
%feature fusion, but it expands the representation space through fusing features from different samples, which can make model training more robust and smoother. 

\vspace{0.1cm}\noindent\textbf{Multi-Label Visual Classification --}
Compared with typical single-label image classification, the problem of multi-label classification \cite{Zhang2014ARO, Liu2021TheET} is more challenging in view of complicated object co-occurrence in the images, which leads to large variations of  visual representations.
%as a multi-label image always contains more than one different objects, which are possibly with small size and large variance.
With the rise of deep networks \cite{he2016deep, Huang2017DenselyCC}, most recent works \cite{Chen2019MultiLabelIR, Yun2021RelabelingIF, Baruch2020AsymmetricLF, Wang2016CNNRNNAU, Chen2018RecurrentAR, zhou2021deep, Wen2021MultilabelIC} tend to address the problem via deep representation learning and a multi-label classification layer to model latent correlation across targets. 
%predict multi-label targets directly.
%Wei \etal \cite{Wei2016HCPAF} proposed a flexible deep CNN framework with a Hypotheses-CNN-Pooling (HCP) operation for multi-label classification, while without using any additional prior information except for label ground-truth.
%Meanwhile, Wang \etal \cite{Wang2016BeyondOP} proposed a random crop pooling (RCP) operation to sample random region in images for deep network training, which was further enhanced with a dynamic weighted Euclidean loss. 
Wang \etal \cite{Wang2017MultilabelIR} developed a recurrent memorized-attention module contained with a spatial transformer layer and a LSTM unit for capturing the correlation between local and global regions.
Chen \etal \cite{Chen2019MultiLabelIR} introduced Graph Convolutional Networks (GCN) into popular deep model to learn the label correlation with an end-to-end trainable way.
Recently, Wen \etal \cite{Wen2021MultilabelIC} proposed a feature and label co-projection (CoP) module to explicitly model the context of multi-label images, inspired by the psychological way of human recognizing multiple objects simultaneously. 
Although those algorithms with pre-trained deep networks can alleviate the problem of multi-label classification task, there are still suffer from lack of sufficient training data to model cross-object dependency. 
Our method can vastly augment more training samples in the feature space with mixed multi-labels to alleviate the challenge as well as better generalization.

%-----------------------------------------------------------------------------------------
\section{Methodology}\label{sec:methods}
In Sec. \ref{subsec:pre}, we firstly introduce some notations used in this paper as well as several  representative mixup style data augmentation methods, \ie the Mixup \cite{zhang2018mixup} and the Manifold Mixup \cite{Verma2019ManifoldMB}. 
Secondly, the proposed ShuffleMix method is presented in Sec. \ref{subsec:shufflemix}.
Finally, we 
%briefly analyze the complexity of the proposed ShuffleMix method.
%Finally, we 
combine our ShuffleMix with the Noisy Injection scheme in Sec. \ref{sec:nosie-mixup}.

\subsection{Preliminaries}\label{subsec:pre}
In supervised single-label classification (\textcolor{black}{\eg general image classification \cite{Krizhevsky2012ImageNetCW,Szegedy2016RethinkingTI,he2016deep} and fine-grained image classification \cite{huang2016part, fu2017look, zheng2019learning, du2020fine, Chang2020TheDI, Shi2019FineGrainedIC, Ding2021APCNNWS}}), given a training data set ${\{(x_i,y_i)\}_{i=1}^N}$, the model $f(x)$, \eg a deep neural network, is usually trained by minimizing the {\em empirical risk} (ERM):
\begin{equation}\label{eq:erm}
    R(f) = \frac{1}{N} \sum_{i=1}^{N} \ell(f(x_i), y_i),
\end{equation}
where $x_i \in \mathbb{R}^{W \times H \times C}$ denotes the $i$-th input image with width $W$, height $H$ and color channels $C$, $y_i$ denotes the corresponding label, and $\ell(\cdot)$ is the loss function, \eg the typical cross entropy loss. 
To improve model generalization, there appears a series of mixup style data augmentation, \eg Mixup \cite{zhang2018mixup} and Manifold Mixup \cite{Verma2019ManifoldMB}.

The vanilla Mixup \cite{zhang2018mixup} method generates a new interpolated image $\Tilde{x}$ and correspondingly semantically interpolated label $\Tilde{y}$ by combining any two training images $(x_a, y_a)$ and $(x_b, y_b)$ with a random trade-off value $\lambda$, which can be written as follows:
\begin{equation}\label{eq:mixup}
    \begin{aligned}
        &\Tilde{x} =  \lambda x_a + (1 - \lambda) x_b, \\
        &\Tilde{y} =  \lambda y_a + (1 - \lambda) y_b,  \\
    \end{aligned}
\end{equation}
where $\lambda \thicksim Beta(\alpha, \alpha)$, for $\alpha \in (0, +\infty)$. When $\alpha$ is set to 1, the value of $\lambda$ is the same as sampling from a uniform distribution $U(0, 1)$.
The generated image $\Tilde{x}$ is used to train the model $f(x)$, supervised by the interpolated label $\Tilde{y}$. 

Recently, the Manifold Mixup \cite{Verma2019ManifoldMB} is proposed to leverage interpolations in deeper hidden layers rather than only in the input space for exploiting manifolds of high-level representations. 
We train deep image classifiers with the Manifold Mixup based data augmentation with the following steps.
\begin{itemize}
\item First, supposing a model $f(x) = f_k \circ g_k(x)$, where $g_k$ denotes the shallow $k$ feature encoding layers and $f_k$ denotes all the remaining ones. The $g_k(x) \in \mathbb{R}^{W' \times H' \times C'}$ represents the hidden representation at layer $k$, where $k$ is randomly selected from a range of eligible layers $\mathcal{S} = \{k\}$ in the whole classification network.
\item Second, the interpolated representation $\Tilde{g_k}(x)$ is generated by performing the vanilla Mixup \cite{zhang2018mixup} on two random hidden representations $g_k(x_a)$ and $g_k(x_b)$, which are encoded from training images $(x_a, y_a)$ and $(x_b, y_b)$ respectively, with a random value $\lambda$, illustrated as follows:
\begin{equation}\label{eq:manifoldmixup}
    \Tilde{g_k}(x) =  \lambda g_k(x_a) + (1 - \lambda) g_k(x_b),
 \end{equation}
where $k \in \mathcal{S}$, $\lambda \thicksim Beta(\alpha, \alpha)$ and the corresponding semantically interpolated label $\Tilde{y} =  \lambda y_a + (1 - \lambda) y_b$. 
\item Third, the generated representation $\Tilde{g_k}(x)$ is used to additionally train the model $f_k(\Tilde{g_k}(x) )$ under supervision of $\Tilde{y}$. 
Note that, the Manifold Mixup \cite{Verma2019ManifoldMB} in the case where $\mathcal{S} = \{0\}$, is equivalent to the vanilla Mixup \cite{zhang2018mixup}.
\end{itemize}

\subsection{Shuffle Mixup of Neural Hidden States}\label{subsec:shufflemix}

As the hidden representations in the middle layers preserving low-dimensional manifolds, feature interpolations of different category samples introduced by the Manifold Mixup \cite{Verma2019ManifoldMB} are enforced to avoid label inconsistency for the same augmented data point with different original sample pairs by flattening their features, which can thus improve model generalization via increasing margins to smoother decision boundaries.
%regularizing features of samples belonging to the same class more tight.    
As mentioned in Sec. \ref{sec:intro}, simple linear operations are not limited to the linear interpolation in the Manifold Mixup \cite{Verma2019ManifoldMB}, which encourage us to use a generalized linear combination to more flexibly extend the capacity of representations constrained by low-dimensional manifolds.
  
%Considering the hidden representations at middle layer of model always highly aligned, the interpolation between two hidden representations of different samples could lead to a misunderstanding of the third part category \cite{Mikolov2013EfficientEO}. Meanwhile, the mixed label is just a linear combination of two categories, which is not enough to represent the third part category. Based on this observation, Manifold Mixup \cite{Verma2019ManifoldMB} may limit the capacity of representation learning to some extent.

Motivated by the efficient ShuffleNet \cite{Zhang2018ShuffleNetAE}, this paper proposes the ShuffleMix method, which favors channel-wise linear operations such as interpolation and shuffle on feature maps.
Specifically, our ShuffleMix can preserve a part of feature channels, while the rest of maps perform the typical interpolation as other mixup methods on two random hidden representations at the same indices.
The advantages of our ShuffleMix are summarized in four folds below.
\begin{itemize}
\item Our ShuffleMix is a generalized scheme to allow the vanilla Mixup and the Manifold Mixup as its special cases.
\item Feature channels randomly selected for mixup can be considered as soft dropout regularization to improve model generalization (see results in Sec. \ref{sec:generalization}).
\item Ratio of the preserved feature maps can control perturbation boundary of the augmented features as an implicit smoothness regularization to improve model robustness against noises (see results in Sec. \ref{sec:robustness}).
\item Feature mixing between different samples can vastly improve representations not only for single-label classification (see Sec. \ref{sec:generalization}), but also for multi-label classification (see Sec. \ref{sec:multi-label}).
\end{itemize}

\textcolor{black}{
Training the model $f(x)$ with our {ShuffleMix} are largely similar to the Manifold Mixup, while the main difference lies in the second step, where our {ShuffleMix} allows not only feature interpolation but also channel-wise shuffle operation on feature maps from hidden layers of neural networks.
Especially, a randomly-generated binary index vector $\bm{m}$ is employed for selecting channel dimensions to be shuffled in-between feature maps of two samples, which can be formulated into the following equation:
\begin{equation}\label{eq:hard_shufflemix}
    \Tilde{g_k}(x) =  (\mathbf{1} - \mathbf{m}) \ \odot \ g_k(x_a) 
     + \mathbf{m} \ \odot \  g_k(x_b), 
\end{equation}
where $\mathbf{m} \in \{0, 1\}^{C'}$ denotes a binary index along the channel dimension for indicating which channel need to be performed with a linear interpolation of channel-wise dropout feature maps, $\mathbf{1}$ is a binary vector filled with all one elements, and $\odot$ denotes channel-wise multiplication. 
}

\textcolor{black}{
Compared with the Manifold Mixup in Eq. (\ref{eq:manifoldmixup}), our proposed ShuffleMix operation in Eq. (\ref{eq:hard_shufflemix}) plays a hard shuffle mixup of hidden states instead of leveraging linear mixup between feature maps of two samples. 
To make our ShuffleMix more flexible, we further extended the hard shuffle mixup to a soft shuffle mixup via replacing a part of randomly-selected feature channels with mixup interpolated features.
The formula can be depicted as follows:
\begin{equation}\label{eq:soft_shufflemix}
\begin{aligned}
    \Tilde{g_k}(x) & =  (\mathbf{1} - \mathbf{m}) \ \odot \ g_k(x_a) \\
    & + \mathbf{m} \ \odot \ \{\lambda g_k(x_a) + (1 - \lambda) g_k(x_b)\}, \\
\end{aligned}
\end{equation}
where $\lambda$ is a random value sampled from the beta distribution $Beta(\alpha, \alpha)$ like the vanilla Mixup \cite{zhang2018mixup}.
We define such an operation in Eq. (\ref{eq:soft_shufflemix}) with $\lambda \thicksim Beta(\alpha, \alpha)$ as soft ShuffleMix, while the operation in Eq. (\ref{eq:hard_shufflemix}) is named as hard ShuffleMix, easily degenerated from soft ShuffleMix with $\lambda \cong 0$. 
Performance comparisons between soft and hard ShuffleMix are shown in the experiment part.
% Unless specifically stated, we treat the soft ShuffleMix as our proposed ShuffleMix method.
Unless particularly stated, the results of our ShuffleMix in the experiments are obtained with the soft ShuffleMix.
}
The binary index $\mathbf{m}$ is actually a $C'$-dimensional vector, whose sum $||\mathbf{m}||$ indicates the number of indexed channels. 
Given a hyper parameter $r \in (0,1]$ for representing the ratio of indexed channels, we can obtain $||\mathbf{m}|| = |r*C'|$. 
Note that, $r$ should be larger than zero, and when $r=1$, our ShuffleMix is equivalent to the Manifold Mixup \cite{Verma2019ManifoldMB}. 
%The value of hyper parameter $r$ is determined by experiments. 
Given $\lambda$ and $r$, we can calculate the semantically interpolated label $\Tilde{y}$ as the following:
%, which tend to move closer to itself.  
\begin{equation}\label{eq:gm2-label}
    \Tilde{y} = (1 - r) y_a + r \{\lambda y_a + (1 - \lambda) y_b\}.
\end{equation}
It is worth mentioning here that with an appropriate $r$, augmented features can always fall into the sample $x_a$'s neighborhood, which can be viewed as implicitly incorporating data perturbation, similar to the Noisy Feature Mixup \cite{Lim2021NoisyFM} (see Sec. \ref{sec:nosie-mixup}).
%We can always make the model learn more information for sample $x_a$ than sample $x_b$ by setting a reasonable value of $r$. This is the main novelty of our {\em Shuffle Mixup of Neural Features from Hidden Layers}.

%The pipeline of our proposed GM2 is described in Algorithm \ref{alg:gm2}.
\begin{algorithm}[t]
    \caption{The pipeline of our ShuffleMix}
    \label{alg:gm2}
    \begin{algorithmic}[1]
    \REQUIRE Dataset $\mathcal{D}_{train}$ and $\mathcal{D}_{test}$, model $f(x)=f_k \circ g_k(x)$, eligible layers $\mathcal{S} = \{k\}$, ratio $r$, hyper-parameter $\alpha$.
    \ENSURE Predicted output $\hat{y}$
    \IF{training}
            \STATE Random sample a batch of training data $\mathcal{B} \subset \mathcal{D}_{train}$
            \FOR{$i = 1$ \textbf{to} $|\mathcal{B}|$}
                \STATE Random generate a value $\lambda$ with $Beta(\alpha, \alpha)$
                \STATE Random select a number k from layers $\mathcal{S}$ for hooking hidden representation
                \STATE Generate random binary index $\mathbf{m}$ with ratio r
                \STATE Obtain hidden representation $g_k(x_i)$ at k layer
                \STATE Generate mixed feature $\Tilde{g_k}(x_i)$ by \textbf{Eq.}(\ref{eq:soft_shufflemix})
                \STATE Generate mixed label $\Tilde{y_i}$ by \textbf{Eq.}(\ref{eq:gm2-label})
                \STATE Obtain the predicted output $\hat{y}$ with $f_k \circ \Tilde{g_k}(x_i)$
            \ENDFOR
            \STATE Compute the empirical risk according to \textbf{Eq.}(\ref{eq:erm}) 
            \STATE Backward to update the parameters of the model     
    \ELSE
        \STATE Random sampling testing data $x \subset \mathcal{D}_{test}$    
        \STATE Compute the predicted output $\hat{y} = f(x)$
    \ENDIF
    \RETURN $\hat{y}$
\end{algorithmic}
\end{algorithm}

%\subsection{Complexity of ShuffleMix}
As shown in Eq. (\ref{eq:hard_shufflemix}) and Eq. (\ref{eq:soft_shufflemix}), the operation $\odot$ can be implemented simply with a way of channel replacement in practice, while the replacement operation is theoretically cost-free. Hence, our ShuffleMix algorithm has the same computational and space complexity
% \footnote{\textcolor{red}{computational complexity?}} 
as that of the Manifold Mixup \cite{Verma2019ManifoldMB} algorithm, which can be also verified by our experiments.
The detailed training processing of our ShuffleMix is presented in Algorithm \ref{alg:gm2}.

\subsection{Combining with Noisy Feature Mixup}\label{sec:nosie-mixup}

The Noisy Feature Mixup (NFM) \cite{Lim2021NoisyFM} was recently proposed to use the Manifold Mixup \cite{Verma2019ManifoldMB} by injecting additive and multiplicative noises into the interpolated representations $\Tilde{g_k}(x)$ in Eq. (\ref{eq:manifoldmixup}) as feature regularization, which can be depicted as:
\begin{equation}\label{eq:nfm}
    \Tilde{g_k}(x)_\text{NFM} = (1 + \delta_{mult} \xi_{k}^{mult}) \odot \Tilde{g_k}(x) + \delta_{add} \xi_{k}^{add},
\end{equation}
where $\delta_{mult}$ and $\delta_{add}$ are pre-defined hyper parameters for leveling additive noise $\xi_{k}^{mult}$ and multiplicative noise $\xi_{k}^{add}$ respectively. 
As mentioned in NFM \cite{Lim2021NoisyFM}, $\xi_{k}^{mult}$ and $\xi_{k}^{add}$ are independent random variables, which are sampled from a normal distribution and a uniform distribution for approximating white and salt \& pepper noises respectively. 
Other training details are kept the same as that in the Manifold Mixup \cite{Verma2019ManifoldMB}.

Motivated by performance gain by the combination of mixup style data augmentation and noise injection, we similarly  train deep representation learning with combining our ShuffleMix and noisy feature injection. 
This method named as ShuffleMix-NFM based on Eq. (\ref{eq:nfm}) replaces the interpolated representation $\Tilde{g_k}(x)$ of the Manifold Mixup \cite{Verma2019ManifoldMB} with those of our ShuffleMix in Eq. (\ref{eq:soft_shufflemix}).
%. Especially note, the main difference of $\Tilde{g_k}(x)$ between  and Manifold Mixup \cite{Verma2019ManifoldMB} can make the model training more robust intuitively. 

As illustrated in Eq. (\ref{eq:soft_shufflemix}) and Eq. (\ref{eq:hard_shufflemix}), the interpolated representation $\Tilde{g_k}(x)$ in our ShuffleMix not only contains interpolated features, but also preserves original ones.
%contains unmixed representations, but also mixed representations. 
Therefore, combination of $\Tilde{g_k}(x)$ in our ShuffleMix and injected noises in Eq. (\ref{eq:nfm}) can achieve superior robustness owing to data perturbation on both raw and mixed representations, rather than only the latter in the manifold mixup, which is supported by experimental results in Sec. \ref{sec:generalization}.
%inherit the benefits from mixup style data augmentation, noise injection and undisturbed data information at the same time. 

%---------------------------------------------------------------------------
\section{Adaptation to Multi-Label Classification}\label{sec:application}
% \textcolor{black}{Ke: treat fine-grained classification and multi-label classification as applications of the proposed shufflemix here.}
% In this section, we introduce our method to three different visual tasks. Firstly, we combine our method with Noisy Feature Mixup for achieving better robustness for input noise injection. Secondly, we introduce our method to fundamental fine-grained classification. Thirdly, we apply our method to multi-label classification with some indispensable modifications. 
\textcolor{black}{To further illustrate the effectiveness of our proposed methods, we 
%treat fine-grained classification and multi-label classification as the applications of our proposed ShuffleMix. We first introduce our ShuffleMix to fundamental fine-grained classification just as a kind of single-label classification, and then 
adapt our method to multi-label classification,}
which was
%is built upon Mixup \cite{zhang2018mixup} data augmentation, which is 
originally designed for single-label image classification.
The challenge in multi-label representation lies in more complicated label dependency, which leads to large variations of visual representations, in comparison with single-label classification.
%recent studies \cite{Yun2021RelabelingIF, Baruch2020AsymmetricLF} have shown that most natural images contain multiple contents and objects in practice, which means an urgent requirement of multi-label classification.
%In comparison, multi-label classification is more complicated and challenging than single-label classification.
%In this section, to demonstrate the generalization of our ShuffleMix, we further introduce how to implement our ShuffleMix method in multi-label image classification task. 
%The aim of multi-label classification \cite{Zhang2018MultilabelIC, Chen2019MultiLabelIR, zhou2021deep} is to predict a series of objects present in a given image, as illustrated in Figure \ref{fig:images3}.
This task commonly train a multi-label classifier $f(\cdot)$ on the training set ${\{(x_i,y_i)\}_{i=1}^N}$ with multi-label binary cross-entropy loss as follows:
\begin{equation}\label{eq:multi-fc} 
    % L = \frac{1}{N} \sum_{i=1}^{N} \sum_{k=1}^{K} \ell (\hat{y_i}^{k}, y_{i}^{k})
    \ell(\cdot) = \sum_{k=1}^{K} - y_{i}^{k}\log(\sigma(\hat{y_i}^{k})) - (1-y_{i}^{k}) \log(1-\sigma(\hat{y_i}^k)),
\end{equation}
where $K$ denotes the number of labels, $y_i\in{\{0,1\}}^K$ denotes the target multi labels, $\hat{y_i}=f(x_i)$ denotes the predicted multi labels, and $\sigma(\cdot)$ denotes the sigmoid function.

Considering the target labels of binary cross-entropy loss in Eq. (\ref{eq:multi-fc}) are always either zero or one, we can't directly apply the ShuffleMix with soft labels $\Tilde{y}$ in Eq. (\ref{eq:gm2-label}) for multi-label classification. 
% \textcolor{red}{as soft labels are usually beneficial to distill knowledge from noisy labels rather than identify true labels\footnote{Ke: I cannot understand the meaning of this sentence.}}. 
% \textcolor{red}{as soft labels are usually beneficial to mitigate label noise rather than identify true labels}. 
For the ShuffleMix data augmentation in Eq. (\ref{eq:soft_shufflemix}), a straightforward solution is to set values of soft labels greater than zero to be one. 
However, such an approximation will introduce external label noises for those subtle label values. 
For focusing on the dominant targets in multi-label data augmentation, we introduce a pre-defined threshold $m$ for soft labels $\Tilde{y}$ as follows:
\begin{equation}\label{eq:threshold} 
    \Tilde{y}_t = \left\{ \begin{array}{rcl}
       1  & \text{if} \ \Tilde{y} \geq m  \\
       0  & \text{otherwise},
    \end{array}\right.
\end{equation}
where $\Tilde{y} \in \mathbb{R}^K$, $\Tilde{y}_t \in {\{0,1\}}^K$.
%, and $m$ is a predefined threshold. 
Therefore, we can employ the adjusted binary labels $\Tilde{y}_t$ to train classification models with the loss function in Eq. (\ref{eq:multi-fc}) for feature augmentation for multi-label classification. The optimal value of $m$ will be determined empirically, whose ablation studies are shown in the experiments.
To our best knowledge, our work is the first attempt to apply mixup augmentation to multi-label classification, and experiment results in Sec. \ref{sec:multi-label} can verify the effectiveness of our ShuffleMix for multi-label classification.

%-----------------------------------------------------------------------------------------
\section{Experiments}\label{sec:exps}

% \textcolor{black}{Ke: It is better to re-organize this section according to the claim in the introduction.  1. Hard vs Soft ShuffleMix; 2. Table III and IV are not so clear. 3. it is better to add visualization of smooth boundaries as Manifold Mixup. 4. do we have an experiment to indicate the capability of dropout style augmentation against overfitting? 5. Robustness aganist noises should not be over-claimed as it is not our main contributions}

In this section, we empirically validate the proposed ShuffleMix algorithm on popular benchmarks of single-label visual classification tasks, \ie
%Firstly, we implement our algorithm on typical 
public benchmarks of general image classification, \ie the CIFAR \cite{krizhevsky2009learning} and the Tiny ImageNet \cite{Le2015TinyIV}, and 
%Secondly, we show the robustness of the ShuffleMix against perturbation noises on CIFAR-100 dataset \cite{krizhevsky2009learning} with different backbones. 
%Thirdly, to verify the generalization of the ShuffleMix, we conduct experiments on several widely used 
three fine-grained datasets including the CUB-200-2011 (CUB) \cite{wah2011caltech}, the Stanford Cars (CAR) \cite{krause20133d} and the FGVC Aircraft (AIR) \cite{maji2013fine}. 
Furthermore, the ShuffleMix is adapted to the multi-label classification task on the Pascal VOC 2007 dataset \cite{Everingham2009ThePV}.
Finally, visualizations of decision boundaries of different methods are compared for better understanding.

\begin{figure}[t]
    \centering
    % \subfigure[General Single-label Images from ImageNet]{ 
    %         \includegraphics[width=0.95\linewidth, trim=0 5 0 0,clip]{figures/Images_1.png}
    %         \label{fig:images1}}\vspace{-0.15cm}
    % \quad
    \subfigure[Single-label samples from the CUB-200-2011]{ 
            \includegraphics[width=0.95\linewidth, trim=0 5 0 0,clip]{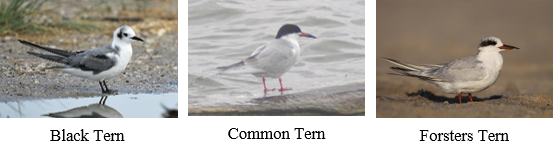}
            \label{fig:images2}}\vspace{-0.15cm}
    \quad
    \subfigure[Multi-label samples from the PASCAL VOC 2007]{ 
            \includegraphics[width=0.95\linewidth, trim=0 5 0 0,clip]{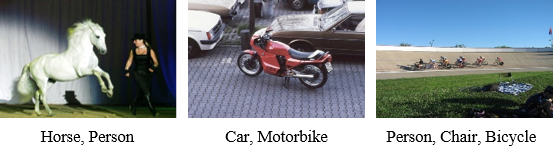}
            \label{fig:images3}}\vspace{-0.15cm}
    \caption{Examples from the CUB-200-2011 \cite{wah2011caltech} and the PASCAL VOC 2007 \cite{Everingham2009ThePV} datasets. The category labels corresponding to the images are displayed below the image. 
    The single-label samples generally just contain one object, while those for multi-label classification usually contains more scene objects.}
    \label{fig:images}%\vspace{-0.2cm}
\end{figure}

%In this section, we give the details of datasets and experimental settings (Section \ref{sec:setting}). We study the performance of ShuffleMix compared with baseline and other representative mixup style data augmentation (Section \ref{sec:generalization}). We verify the robustness of ShuffleMix when testing with noisy data, as well as demonstrate the effectiveness of its extension ShuffleMix-NFM (Section \ref{sec:robustness}). 

\subsection{Datasets and Settings}\label{sec:setting}

\vspace{0.1cm}\noindent\textbf{Datasets --}
We conduct experiments on three widely used datasets: the CIFAR-10 \cite{krizhevsky2009learning}, the CIFAR-100 \cite{krizhevsky2009learning} and the Tiny ImageNet \cite{Le2015TinyIV} for general image classification. 
Specifically, the CIFAR-10 and the CIFAR-100 datasets both contain 50,000 training images and 10,000 test images, belonging to 10 and 100 semantic object classes respectively, while the Tiny ImageNet consists of 100,000 training images of 200 classes and 10,000 test images. 
Note that, both CIFAR and Tiny ImageNet datasets have a balanced distribution for both training and test sets.
For the fine-grained classification task, we conduct experiments on three widely used fine-grained datasets, namely the CUB-200-2011 (CUB) \cite{wah2011caltech}, the Stanford Cars (CAR) \cite{krause20133d} and the FGVC Aircraft (AIR) \cite{maji2013fine}. 
The details of datasets are provided in Table \ref{table:fine-datasets}, where we only use the category labels as supervision signals without any additional prior information.

\begin{table}[t]
\centering
\caption{Statistics of three fine-grained datasets.}
\label{table:fine-datasets}
\begin{tabular}{@{}lccc@{}}
\toprule
Dataset       & Classes   & Training data  & Testing data \\ \midrule
CUB \cite{wah2011caltech}  & 200      & 5,994      & 5,794  \\
CAR \cite{krause20133d}  & 196       & 8,144      & 8,041  \\
AIR \cite{maji2013fine}   & 100      & 6,667      & 3,333  \\ \bottomrule
\end{tabular}
\end{table}

\vspace{0.1cm}\noindent\textbf{Comparative Methods --}
To better evaluate the generalization of our method, we conduct experiments by employing three varieties of the ResNet \cite{he2016deep}, \ie PreActResNet18, PreActResNet34 and Wide-PreActResNet18-2, which are also adopted in our direct competitors -- the Manifold mixup \cite{Verma2019ManifoldMB} and the NFM \cite{Lim2021NoisyFM}. 
Following \cite{Verma2019ManifoldMB,Lim2021NoisyFM}, we use the three networks just with ERM as the baseline.
We compare our method with typical Dropout \cite{Srivastava2014DropoutAS} regularization and representative mixup style data augmentation methods, \ie Input Mixup \cite{zhang2018mixup} and Manifold Mixup \cite{Verma2019ManifoldMB}. 
On evaluation about robustness against noises, the state-of-the art NFM \cite{Lim2021NoisyFM} is competed with the ShuffleMix method.

\vspace{0.1cm}\noindent\textbf{Implementation Details --}
In all our experiments for general image classification, we adopt stochastic gradient descent (SGD) with momentum of 0.9 and a weight decay of $5\times10^{-4}$ for training models.
Following implementation details in \cite{Lim2021NoisyFM}, all our experiments are trained for 200 epochs with a batch size of 128 and a step-wise learning rate decay.
The initial learning rate is set as 0.1 and then decayed with a factor of 0.1 at the 100-th epoch, the 150-th epoch and the 180-th epoch, respectively.
In Eq. (\ref{eq:soft_shufflemix}), there are three hyper parameters, \ie $\alpha$ for beta distribution, $r$ for the ratio of indexed channels for mixup and $\mathcal{S}$ for the range of eligible layers. 
Unless particularly stated, we empirically set $\alpha=1$, $r=0.5$ and $\mathcal{S}=\{0,1,2,3,4\}$ in our experiments.
For fine-grained image classification, we prepare data and implement model training by following the same setting in \cite{Huang2021SnapMixSP, Chang2020TheDI}. The hyper-parameters of the ShuffleMix are setting as same as that in general image classification. For fair comparison, we keep the same training setting for all fine-grained experiments. 
For the evaluation on model robustness, there are another two hyper parameters $\delta_{add}$ and $\delta_{mult}$ for controlling the level of additive noise and multiplication noise, respectively. 
We set $\delta_{add}=0.2$ and $\delta_{mult}=0.4$ in the experiments by following the same setting as \cite{Lim2021NoisyFM}.
To compare with Dropout regularization, we apply Dropout between the final FC layer and feature extractor of each network with a dropping rate of 0.2. The parameters in Mixup and Manifold Mixup are following that in \cite{zhang2018mixup, Verma2019ManifoldMB}. 
We implement and train the models in all experiments with the Pytorch \cite{Paszke2019PyTorchAI} library based on a single NVIDIA 1080ti GPU.

% \usepackage{booktabs}
\begin{table}[t]
\centering
\caption{Comparative evaluation with classification accuracy using three backbone networks on the CIFAR-10 and the CIFAR-100. Standard deviations are obtained over three trails.}
\label{tab:plain} %\vspace{-0.3cm}
\begin{tabular}{@{}lcc@{}}
\toprule
PreActResNet18 & CIFAR-10 (\%)                    & CIFAR-100 (\%)      \\ \midrule
Baseline       & 94.69 $\pm$ 0.172                & 76.12 $\pm$ 0.180                \\
Dropout \cite{Srivastava2014DropoutAS} & 94.63 $\pm$ 0.102       & 76.17 $\pm$ 0.195   \\
Input Mixup \cite{zhang2018mixup}    & 95.67 $\pm$ 0.140                & 79.32 $\pm$ 0.302               \\
Manifold Mixup \cite{Verma2019ManifoldMB} & 95.66 $\pm$ 0.066                & 79.33 $\pm$ 0.249               \\
ShuffleMix (Ours)            & \textbf{95.78} $\pm$ 0.154       & \textbf{80.02} $\pm$ 0.163        \\ \midrule
PreActResNet34 & \multicolumn{1}{l}{} & \multicolumn{1}{l}{} \\ \midrule
Baseline       & 94.74 $\pm$ 0.125               & 76.36 $\pm$ 0.189                \\
Dropout \cite{Srivastava2014DropoutAS}   & 94.72 $\pm$	0.131          &  76.17 $\pm$	0.152             \\
Input Mixup \cite{zhang2018mixup}    & 95.78 $\pm$ 0.103               & 79.54 $\pm$ 0.323                \\
Manifold Mixup \cite{Verma2019ManifoldMB} & 95.57 $\pm$ 0.155               & 79.87 $\pm$ 0.083                \\
ShuffleMix (Ours)             & \textbf{95.78} $\pm$ 0.097      & \textbf{80.53} $\pm$ 0.118       \\ \midrule
Wide-PreActResNet18-2 & \multicolumn{1}{l}{} & \multicolumn{1}{l}{} \\ \midrule
Baseline            & 94.95 $\pm$ 0.040           & 76.90 $\pm$ 0.099           \\
Dropout \cite{Srivastava2014DropoutAS}  & 94.87 $\pm$	0.167        & 77.09 $\pm$	0.142   \\
Input Mixup \cite{zhang2018mixup}         & 96.07 $\pm$ 0.019           & 81.07 $\pm$ 0.039           \\
Manifold Mixup \cite{Verma2019ManifoldMB}      & 95.82 $\pm$ 0.025           & 80.46 $\pm$ 0.311           \\
ShuffleMix (Ours)                  & \textbf{96.36} $\pm$ 0.037  & \textbf{81.96} $\pm$ 0.159  \\ \bottomrule
\end{tabular}
\end{table}

\subsection{Evaluation on Single-Label Image Classification}\label{sec:generalization}

\vspace{0.1cm}\noindent\textbf{Evaluation on General Image Classification --} 
As mentioned in Sec. \ref{sec:setting}, we conduct experiments on the CIFAR-10 and the CIFAR-100 with PreActResNet18, PreActResNet34 and Wide-PreActResNet18-2 backbone networks, respectively. 
Comparative results are reported in Table \ref{tab:plain}, where all methods are under the identical experimental settings.
%mentioned are training and testing in the same environment and setting.
Evidently, the ShuffleMix method can consistently achieve the superior accuracy to comparative mixup style data augmentation methods as well as the baseline, with three backbone networks on both datasets. 
Note that, our method outperforms its direct competitor -- the Manifold Mixup, which is limited to linear interpolations of hidden representations.
Superior performance of the ShuffleMix can be explained by the only difference between both methods lying in the introduction of a more flexible linear combinational operation, which thus can demonstrate our claim. 
%outperforms the Baseline by 5.06\% and outperform the Manifold Mixup \cite{Verma2019ManifoldMB} by 1.5\% on CIFAR-100 with Wide-PreActResNet18-2, which can strongly verify the effectiveness of our method. 
%We observe that the performance of our method on the CIFAR-100 is always better than those on the CIFAR-10.
%Such a phenomenon is caused by data sparsity of two datasets, \ie training images of each class in the CIFAR-10 has ten times more than those in the CIFAR-100. 
%Therefore, the performance improvement in CIFAR-10 will be limited.
%We observe that results of Manifold Mixup \cite{Verma2019ManifoldMB} cannot be are not much or worse than Input Mixup \cite{zhang2018mixup}.
%Also notice that the results of Manifold Mixup \cite{Verma2019ManifoldMB} in our setting are not much or worse than Input Mixup \cite{zhang2018mixup}. After analysis, we thought 200 epochs maybe not enough to train Manifold Mixup \cite{Verma2019ManifoldMB} for get better results. This can further demonstrates the effectiveness of our method.

\vspace{0.1cm}\noindent\textbf{Evaluation on Large-Scale Data --} 
To verify generalization of our method on large-scale data, another experiment on the Tiny ImageNet is conducted with using PreActResNet18, PreActResNet34 and Wide-PreActResNet18-2 backbone networks.
%duo to the limitation of computation resource. 
Results of competing methods are compared in Table \ref{tab:tiny_imagenet}, where we can find out that the proposed ShuffleMix can outperform the Baseline by 2.78\%, 4.56\ and 3.04\% on top-1 accuracy when using PreActResNet18, PreActResNet34 and Wide-PreActResNet18-2 backbones, respectively. 
Meanwhile, the result of our ShuffleMix can beat the Manifold Mixup \cite{Verma2019ManifoldMB} by 1.17\%, 0.8\% and 1.66\% on top-1 accuracy when using PreActResNet18, PreActResNet34 and Wide-PreActResNet18-2 backbone, respectively.
Similar results can also be observed when using top-5 accuracy as performance metric. 
Experiment results on the Tiny ImageNet again verify the effectiveness of our method on large-scale data. 

\begin{table}[t]
\centering
\caption{Comparative evaluation with classification accuracy using three backbone networks on the Tiny ImageNet.}
\label{tab:tiny_imagenet}%\vspace{-0.2cm}
\begin{tabular}{@{}lcc@{}}
\toprule
PreActResNet18 & top-1 (\%)      & top-5 (\%)     \\ \midrule
Baseline         & 62.64          & 83.05         \\
Dropout\cite{Srivastava2014DropoutAS}  & 62.46          & 82.64  \\
Input Mixup \cite{zhang2018mixup}      & 64.47          & 84.02         \\
Manifold Mixup \cite{Verma2019ManifoldMB}    & 64.25          & 84.38         \\
ShuffleMix (Ours)              & \textbf{65.42} & \textbf{85.10} \\ \midrule
PreActResNet34 & \multicolumn{1}{l}{} & \multicolumn{1}{l}{}     \\ \midrule
Baseline         & 62.06          & 82.83         \\
Dropout \cite{Srivastava2014DropoutAS}  & 61.59	        & 81.50\\
Input Mixup \cite{zhang2018mixup}       & 64.39         & 84.59         \\
Manifold Mixup \cite{Verma2019ManifoldMB}    & 65.81          & 85.62         \\
ShuffleMix (Ours)              & \textbf{66.61} & \textbf{85.84} \\  \midrule
Wide-PreActResNet18-2 & \multicolumn{1}{l}{} & \multicolumn{1}{l}{}     \\ \midrule
Baseline         & 64.76          & 83.40         \\
Dropout \cite{Srivastava2014DropoutAS} &  64.40	& 84.74 \\
Input Mixup \cite{zhang2018mixup}      & 65.24          & 84.15         \\
Manifold Mixup \cite{Verma2019ManifoldMB}    & 66.14          & 84.86         \\
ShuffleMix (Ours)              & \textbf{67.80} & \textbf{85.76} \\ \bottomrule
\end{tabular}
\end{table}

\vspace{0.1cm}\noindent\textbf{Evaluation on Fine-Grained Image Classification --} 
The accuracy results on three fine-grained datasets are reported in Table \ref{tab:fine-grained}, where we employ two different backbones, \ie ResNet34 \cite{he2016deep} and ResNet50 \cite{he2016deep}. 
Experiments results demonstrate that the ShuffleMix can also better improve the classification performance than Mixup \cite{zhang2018mixup} and Manifold Mixup\cite{Verma2019ManifoldMB} on fine-grained datasets, which indicates that the proposed ShuffleMix can also benefit the learning of fine-grained classification.
Especially, our ShuffleMix with ResNet34 backbone on AIR \cite{maji2013fine} dataset can outperform the baseline by near 2\%.

\begin{table}[t]
\centering
\caption{Comparison with classification accuracy using ResNet34 and ResNet50 backbones on the CUB-200-2011 (CUB), the Stanford Cars (CAR), and the FGVC-Aircraft (AIR) benchmarks.}
\label{tab:fine-grained}%\vspace{-0.2cm}
\begin{tabular}{@{}lccc@{}}
\toprule
ResNet34       & CUB (\%)        & CAR (\%)        & AIR (\%)         \\ \midrule
Baseline       & 84.83 & 91.94 & 89.98 \\
Input Mixup \cite{zhang2018mixup}    & 84.74 & 92.76 & 90.49 \\
Manifold Mixup\cite{Verma2019ManifoldMB} & 85.24 & 92.91 & 90.70 \\
ShuffleMix (Ours)          & \textbf{85.90} & \textbf{93.33} & \textbf{91.96} \\  \midrule
ResNet50 & \multicolumn{1}{l}{} & \multicolumn{1}{l}{} & \multicolumn{1}{l}{} \\ \midrule
Baseline       & 85.46          & 92.89          & 90.97          \\
Input Mixup \cite{zhang2018mixup}    & 85.98          & 93.35          & 90.94          \\
Manifold Mixup  \cite{Verma2019ManifoldMB} & 86.29          & 93.88          & 92.08          \\
ShuffleMix (Ours)       & \textbf{87.00} & \textbf{94.12} & \textbf{92.41} \\\bottomrule
\end{tabular}
\end{table}

\begin{figure*}[htbp]
    \centering
    \subfigure[CUB]{ 
            \includegraphics[width=0.3\linewidth, trim=15 5 55 50,clip]{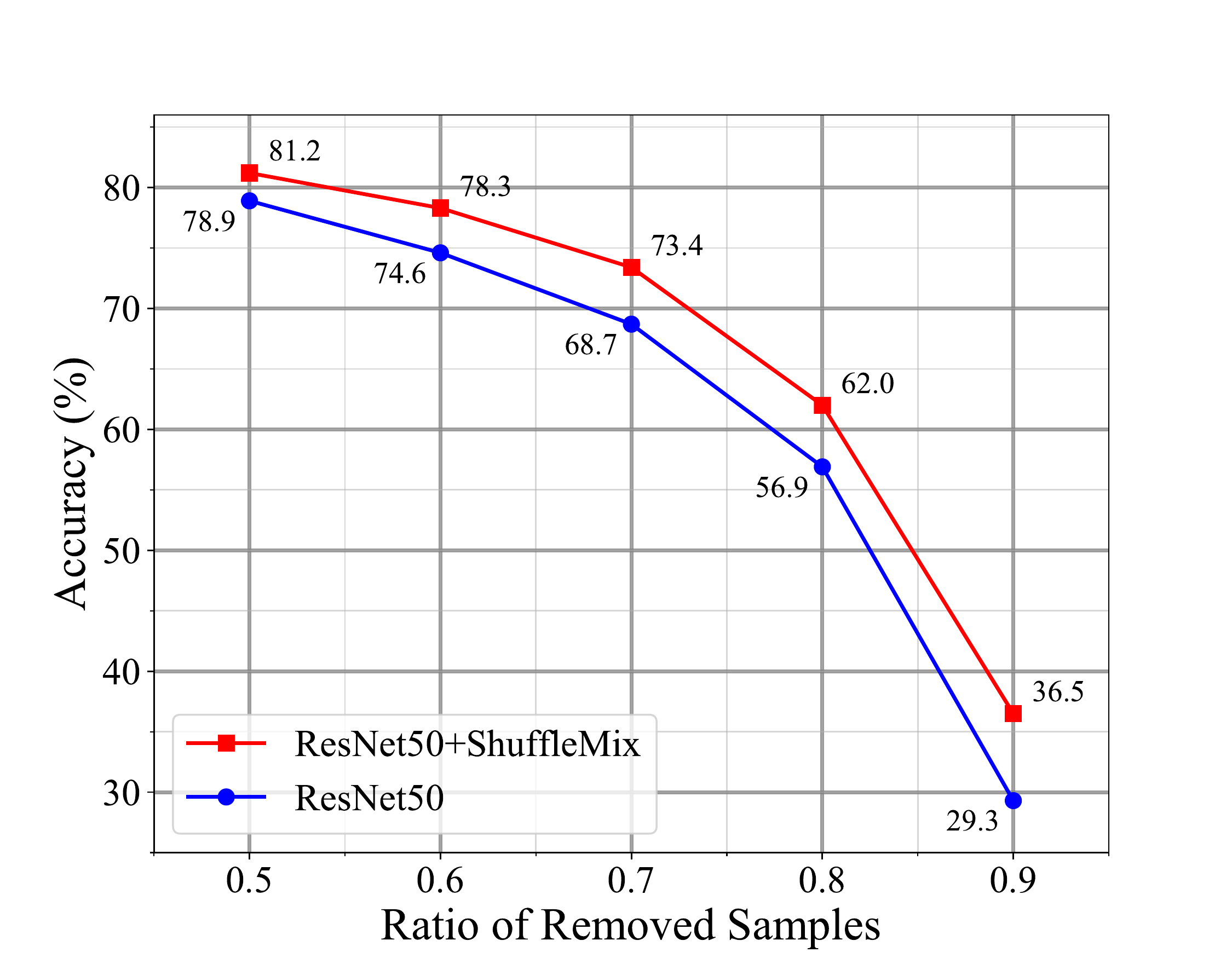}
            \label{fig:sparse_cub}}
    \quad
    \subfigure[CAR]{ 
            \includegraphics[width=0.3\linewidth, trim=15 5 55 50,clip]{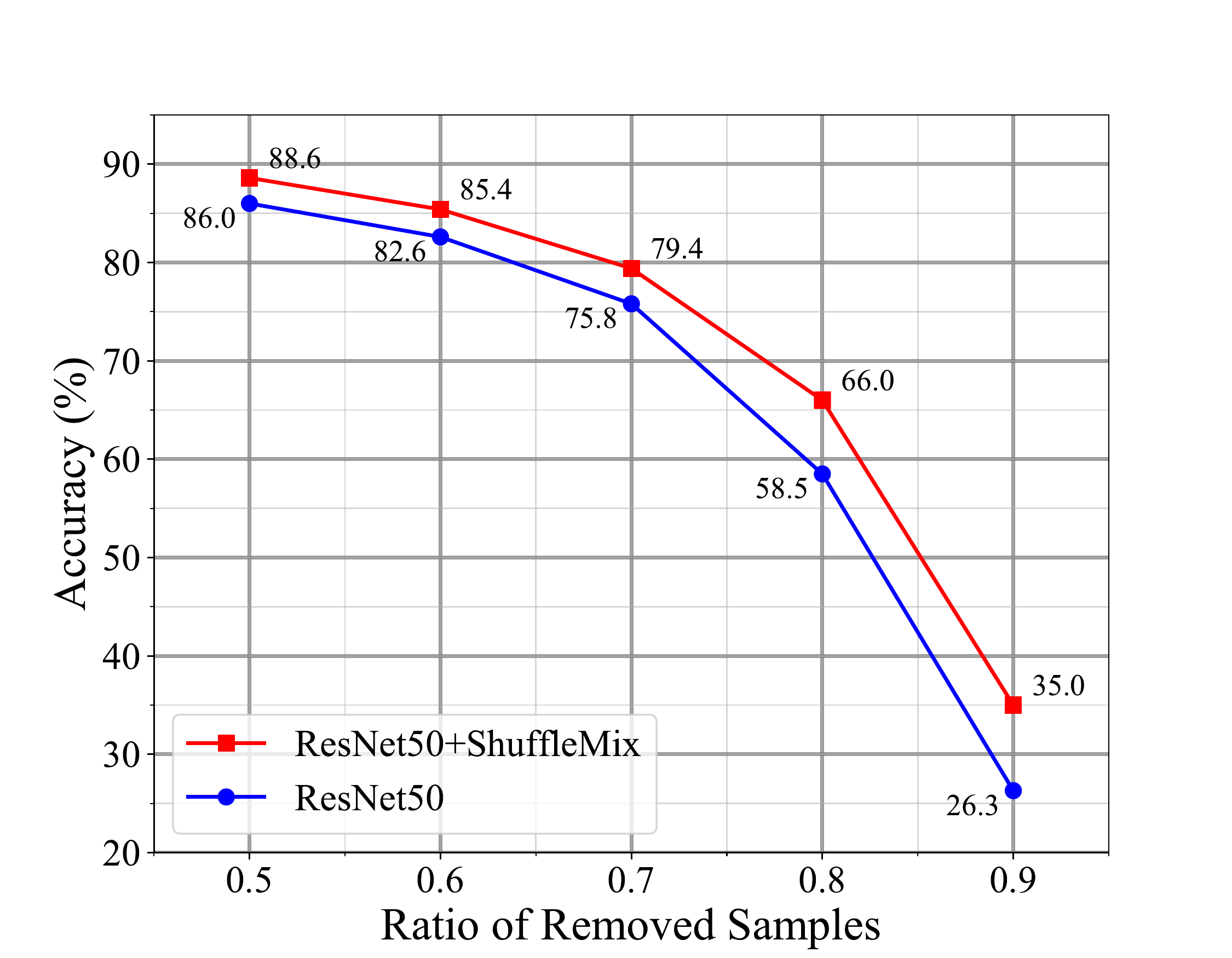}
            \label{fig:sparse_car}}
    \subfigure[AIR]{ 
            \includegraphics[width=0.3\linewidth, trim=15 5 55 50,clip]{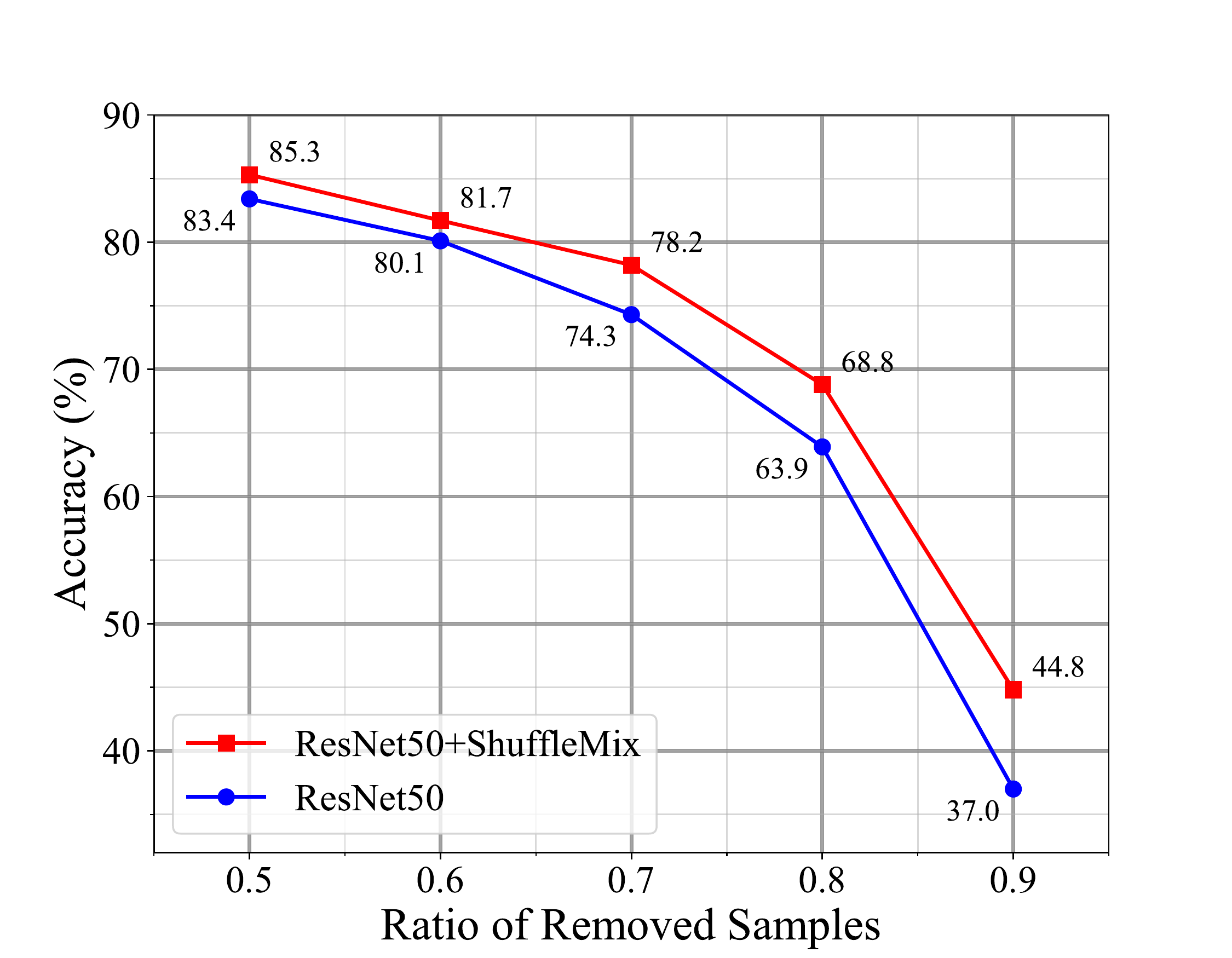}
            \label{fig:sparse_air}}    
    \caption{Classification performance on different fine-grained datasets under varying sample size of each class. (Best viewed in colors)}
    \label{fig:sparse_data}%\vspace{-0.2cm}
\end{figure*}

\begin{table*}[t]
\centering
\caption{Robustness of comparative methods based on PreActResNet18 and Wide-PreActResNet18-2 backbones, w.r.t. white noise ($\delta$) and salt and pepper ($\gamma$) perturbations on the CIFAR-100. The best results are marked in bold, and the second best results are marked in lavender background. Results are averaged over three trials.}
\label{tab:robustness1}%\vspace{-0.2cm}
\setlength{\tabcolsep}{1.2mm}
% \resizebox{\linewidth}{!}{}
\begin{tabular}{@{}lc|ccc|ccc@{}}
\toprule
\multirow{2}{*}{Methods} & \multirow{2}{*}{Clean (\%)} & \multicolumn{3}{c|}{$\delta$ (\%)}                    & \multicolumn{3}{c}{$\gamma$ (\%)}                     \\ \cmidrule(l){3-8} 
                      &                   & 0.1            & 0.2            & 0.3            & 0.02           & 0.04           & 0.1            \\ \midrule
 PreActResNet18 \\                       
\midrule
Baseline                          & 76.12                       & 56.90          & 29.20          & 14.07          & 59.04          & 41.56          & 16.07          \\
Input Mixup \cite{zhang2018mixup}                       & 79.32                       & 65.66          & 39.77          & 19.37          & 58.80          & 40.00          & 15.39          \\
Manifold Mixup \cite{Verma2019ManifoldMB}              & 79.33                       & 63.98          & 33.64          & 15.42          & 63.23          & 45.71          & 15.33          \\
ShuffleMix (Ours)                     & \textbf{80.02}              & {67.03}      & {43.05}          & {21.68}    & {\cellcolor{Lavender!100}70.72}    & {\cellcolor{Lavender!100}60.88}    & {\cellcolor{Lavender!100}35.96}    \\ \midrule
NFM \cite{Lim2021NoisyFM}          & 79.15                       & {\cellcolor{Lavender!100}75.07}  & {\cellcolor{Lavender!100}57.22}     & {\cellcolor{Lavender!100}33.84}          & 70.20          & 59.87          & 31.46          \\
ShuffleMix-NFM (Ours)                 & {\cellcolor{Lavender!100}79.67}                 & \textbf{76.45} & \textbf{61.79} & \textbf{40.45} & \textbf{72.94} & \textbf{66.12} & \textbf{44.42} \\ \midrule \midrule

% \multirow{2}{*}{Wide-PreActResNet18-2} & \multirow{2}{*}{} & \multicolumn{3}{c|}{$\delta$}                    & \multicolumn{3}{c}{$\gamma$}                     \\ \cmidrule(l){3-8} 
                                    %   &                             & 0.1            & 0.2            & 0.3            & 0.02           & 0.04           & 0.1            \\ \midrule
Wide-PreActResNet18-2 & \multicolumn{1}{l}{} & \multicolumn{1}{l}{} & \multicolumn{1}{l}{} & \multicolumn{1}{l}{} & \multicolumn{1}{l}{} & \multicolumn{1}{l}{} & \multicolumn{1}{l}{}     \\ \midrule
Baseline                               & 76.90                       & 56.44          & 27.07          & 12.73          & 58.50          & 39.99          & 14.89          \\
Input Mixup \cite{zhang2018mixup}                            & 81.07                       & 67.88          & 42.53          & 22.25          & 62.32          & 42.58          & 15.65          \\
Manifold Mixup \cite{Verma2019ManifoldMB}                          & 80.46                       & 63.40          & 30.35          & 13.48          & 63.09          & 44.03          & 15.46          \\
ShuffleMix (Ours)                          & \textbf{81.96}              & 70.09          & 43.74          & 21.03          & 70.81          & 58.29          & 31.83          \\ \midrule
NFM \cite{Lim2021NoisyFM}              & 81.20                       & {\cellcolor{Lavender!100}78.03}    & {\cellcolor{Lavender!100}61.73}    & {\cellcolor{Lavender!100}37.05}    & {\cellcolor{Lavender!100}73.02}    & {\cellcolor{Lavender!100}63.16}    & {\cellcolor{Lavender!100}33.60}    \\
ShuffleMix-NFM (Ours)                      & {\cellcolor{Lavender!100}81.58}                 & \textbf{78.67} & \textbf{64.42} & \textbf{42.57} & \textbf{75.83} & \textbf{69.20} & \textbf{48.77} \\ \bottomrule
\end{tabular}
\end{table*}

\subsection{Robustness Evaluation}\label{sec:robustness}

\begin{figure*}[t]
    \centering
    \subfigure{ 
            \includegraphics[width=0.4\linewidth, trim=5 0 60 50,clip]{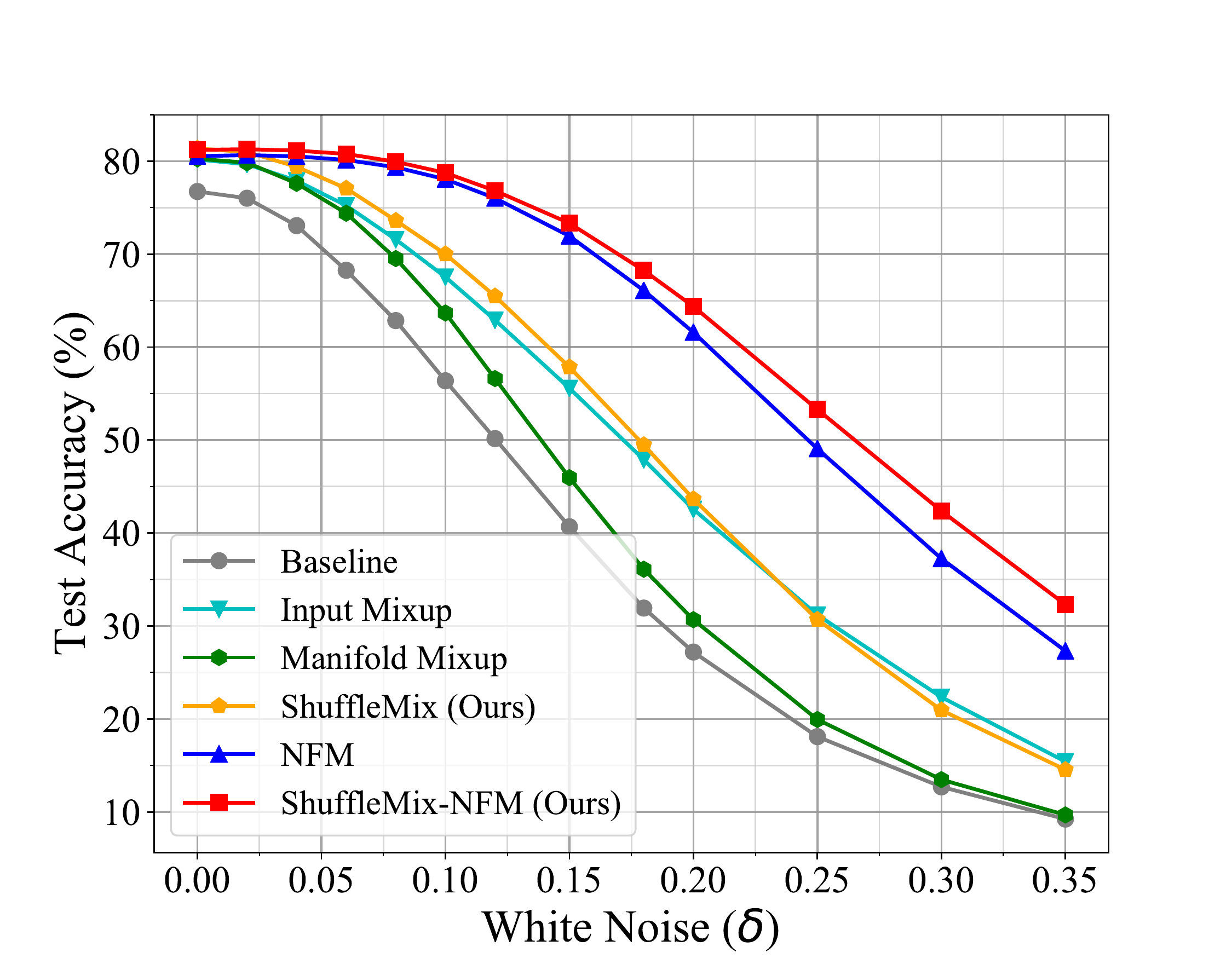}
            \label{fig:white_noise}}
    \quad
    \subfigure{
            \includegraphics[width=0.4\linewidth, trim=5 0 60 50,clip]{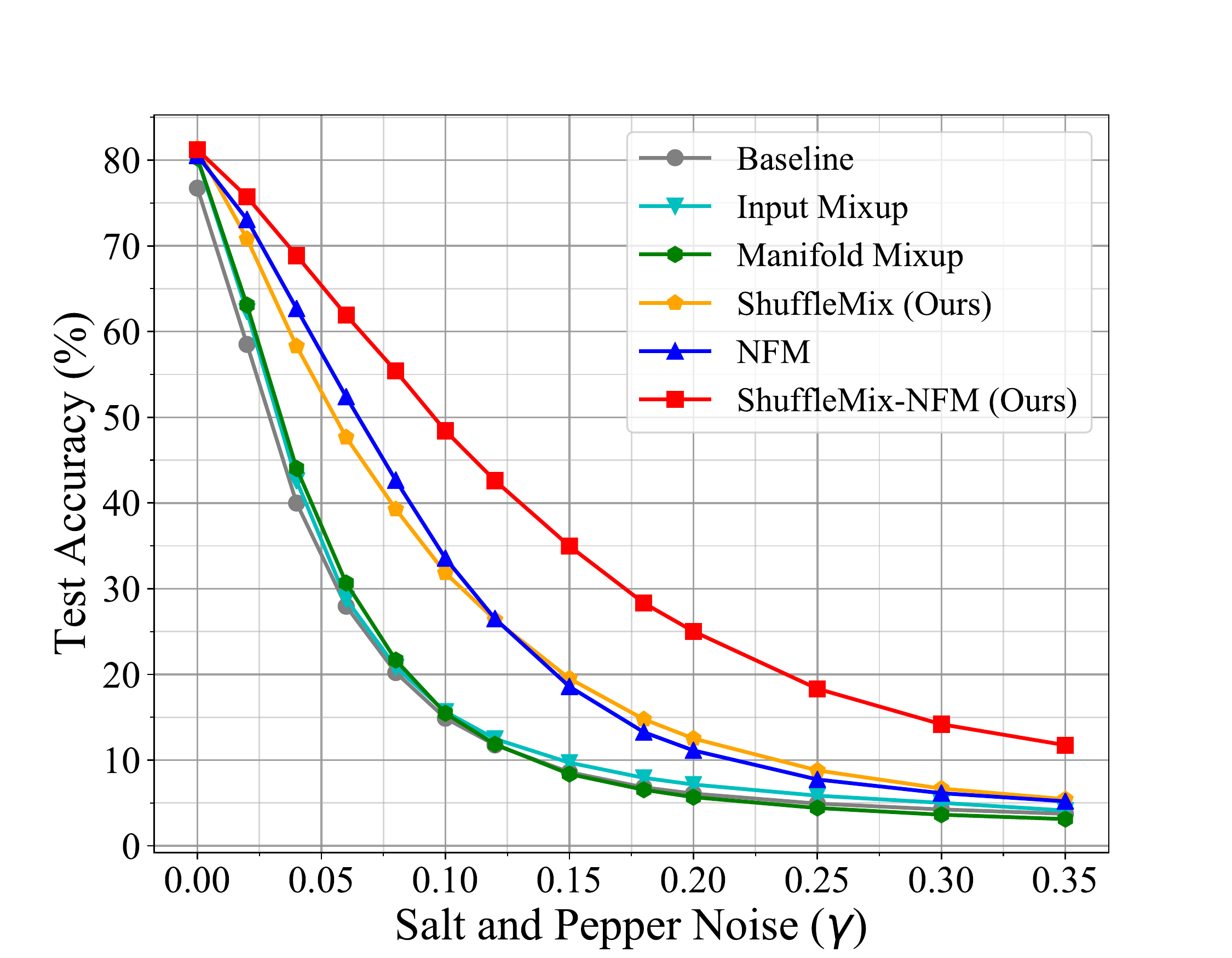}
            \label{fig:sp_noise}}
    \caption{Robustness of comparative methods based on the Wide-PreActResNet18-2 backbone on the CIFAR-100 with classification accuracy. We repeat the experiments for five times to generate the mean accuracy. (Best viewed in color)}
    \label{fig:noise}%\vspace{-0.2cm}
\end{figure*}

\vspace{0.1cm}\noindent\textbf{Robustness Against Noises --}
To verify the robustness of our method ShuffleMix and its extension ShuffleMix-NFM, experiments with different input perturbations, \ie white noise or salt and pepper noise, are conducted. 
Specifically, on the CIFAR-100, the Baseline and other mixup style data augmentation methods, \ie Input Mixup \cite{zhang2018mixup}, Manifold Mixup \cite{Verma2019ManifoldMB} and NFM \cite{Lim2021NoisyFM} are compared with our methods with the same settings mentioned in Sec. \ref{sec:setting}.
Results shown in Tables \ref{tab:robustness1} are based on the backbones PreActResNet18 and Wide-PreActResNet18-2, respectively. 
Results with training models with noise-free data are reported in the first block of both tables, while the Clean indicates that the test data without any perturbation.
$\delta$ and $\gamma$ are used to control the level of white noise and salt \& pepper noise.
Evidently, as illustrated in both tables, it is found out that the ShuffleMix can consistently achieve the best performance on the Clean testing data and also perform more robustly to two types of noises. 
For a fair comparison with the state-of-the-art NFM method \cite{Lim2021NoisyFM}, we train the ShuffleMix-NFM
with the same settings as the NFM, whose results are illustrated in the bottom block of  Table \ref{tab:robustness1} with two different backbones on the CIFAR-100.
The proposed ShuffleMix-NFM can further improve the robustness of models than the ShuffleMix and the  NFM \cite{Lim2021NoisyFM} with significant margins. 
More results about robustness against input perturbations on the Wide-PreActResNet18-2 are illustrated in Fig. \ref{fig:noise}.
The results here can support our claim about improving model robustness via allowing data perturbation in feature space as an implicit regularization.

%Our extension , which are trained with noisy feature, can further improve the robustness of models than the ShuffleMix and even more than NFM \cite{Lim2021NoisyFM} when with salt and pepper noise perturbations. This shows that our method can not only improve the generalization, but also be robust to noise perturbations. 

\vspace{0.1cm}\noindent\textbf{Robustness Against Data Sparsity --}
We further apply the ShuffleMix with varying sparse training samples to verify the robustness against sparse data distribution. 
As both training and test sets are near balanced distributed of datasets in Table \ref{table:fine-datasets}, we set the sparsity of training sets with a limited size of samples for each class (\ie removing 50\% to 90\% of training samples), whereas the whole validation set is adopted in our experiments. 
As shown in Fig. \ref{fig:sparse_data}, our ShuffleMix can always achieve superior results under varying sparsity on different fine-grained datasets, which show our ShuffleMix is more robust against sparse data.

% \usepackage{booktabs}
\begin{table}[t]
\centering
\caption{Classification accuracy of hard and soft ShuffleMix for varying $r$ on the CIFAR-100 with PreActResNet18 backbone.}
\label{tab:shufflemix_ablation}%\vspace{-0.2cm}
\setlength{\tabcolsep}{1mm}
\begin{tabular}{@{}l|ccccccc@{}}
\toprule
ratio $r$              & 0.125 & 0.25  & 0.375 & 0.5           & 0.625          & 0.75  & 0.875 \\ \midrule
\begin{tabular}[c]{@{}l@{}}hard ShuffleMix \\ $\lambda \cong 0$\end{tabular} & 78.90  & 79.05 & \textbf{79.92} & 78.87 & {79.83} & 79.45  & 78.81  \\  \midrule
\begin{tabular}[c]{@{}l@{}}soft ShuffleMix \\ $\lambda \thicksim Beta(1, 1)$\end{tabular} & 78.49 & 78.82 & 79.38 & \textbf{80.02} & 79.63          & 79.56 & 79.46 \\ \bottomrule
\end{tabular}
\end{table}

% % Please add the following required packages to your document preamble:
% % \usepackage{booktabs}
% \begin{table}[t]
% \centering
% \caption{Effects of different $\alpha$ in the proposed ShuffleMix on the CIFAR-100 with PreActResNet18 backbone. The best results in every column are marked underline, and the whole best is marked in bold.}
% \label{tab:alpha}%\vspace{-0.2cm}
% \setlength{\tabcolsep}{2mm}
% \begin{tabular}{@{}lccc@{}}
% \toprule
% $\alpha$ & Input Mixup \cite{zhang2018mixup} & Manifold Mixup \cite{Verma2019ManifoldMB} & ShuffleMix (Ours)                   \\ \midrule
% 0.5      & 79.32       & 79.18          & {\ul 79.75}          \\
% 1        & 79.32       & 79.33          & {\ul \textbf{79.76}} \\
% 2        & 77.95       & 79.26          & {\ul 79.60}          \\
% 4        & 76.50       & 78.42          & {\ul 79.34}          \\
% 8        & 75.33       & 77.75          & {\ul 79.22}          \\ \bottomrule
% \end{tabular}
% \end{table}

% \usepackage{booktabs}
\begin{table}[t]
\centering
\caption{Effects of different $\alpha$ in the proposed ShuffleMix on the CIFAR-100 with PreActResNet18 backbone. The best results in every column are marked in bold.}
\label{tab:alpha}%\vspace{-0.2cm}
\begin{tabular}{@{}l|ccccc@{}}
\toprule
$\alpha$             & 0.5   & 1              & 2     & 4     & 8     \\ \midrule
Input Mixup \cite{zhang2018mixup} & 79.32 & 79.32          & 77.95 & 76.50  & 75.33 \\
Manifold Mixup \cite{Verma2019ManifoldMB} & 79.18 & 79.33          & 79.26 & 78.42 & 77.75 \\
ShuffleMix (Ours) & \textbf{79.75} & \textbf{80.02} & \textbf{79.60} & \textbf{79.34} & \textbf{79.22} \\ \bottomrule
\end{tabular}
\end{table}

%\vspace{0.1cm}\noindent\textbf
\subsection{Ablation Studies}
As aforementioned, all the experiments are conducted on our soft ShuffleMix method by using the same hyper parameters. 
To further compare the difference between our soft ShuffleMix in Eq. (\ref{eq:soft_shufflemix}) and hard ShuffleMix in Eq. (\ref{eq:hard_shufflemix}), 
we jointly conduct ablation studies about the hyper parameter $r$ for controlling the ratio of indexed channels for shuffle mixup on model generalization, with our hard and soft ShuffleMix methods, respectively.
Experiment results are presented in \ref{tab:shufflemix_ablation}, where the hyper parameter $\alpha$ in soft ShuffleMix is set as 1, and the $\lambda \cong 0$ is fixed for hard ShuffleMix.
It is observed that when $r=0.5$ in our soft ShuffleMix and $r=0.375$ in our hard ShuffleMix, we can achieve the best performance, which both suggests an appropriate $r$ for good performance, while the remaining options of $r$ can still perform better than comparative methods in Table \ref{tab:plain}.
Specifically, the result of $r=0.5$ in soft ShuffleMix shows that we can get the best performance when we choose half of random indexed channels to do mixup interpolation, which may be benefited from soft dropout via a generalized linear combination of feature maps.
Similarly, the best results in hard ShuffleMix can be achieved when we perform channel-wise shuffle operation on feature maps with more or less half of random indexed channels, \eg $r=0.375$ and $r=0.625$. 
Such performance comparisons again show that our soft ShuffleMix is more flexible than hard ShuffleMix.
% Similarly, the best results can be achieved when the range of eligible layers $\mathcal{S}$ is set as $\{0,1,2,3\}$ in the soft ShuffleMix and $\{0,1,2,3,4\}$ in the hard ShuffleMix.
% It is worth pointing out that when $r=1$ in the proposed soft ShuffleMix is equivariant to the Manifold Mixup, and results in Table \ref{tab:ratio} show that superior performance can be consistently achieved with other value for $r$, which further verify superiority of the ShuffleMix to existing Manifold Mixup \cite{Verma2019ManifoldMB}.
Effects of hyper parameter $\alpha$ are also investigated in Table \ref{tab:alpha} with $r=0.5$ for our soft ShuffleMix, whose $\lambda \thicksim Beta(\alpha, \alpha)$. 
We observe that the variation of $\alpha$ can make very marginal effects on our method, while the other comparative data augmentation methods are sensitive to the value of $\alpha$.
We conclude that in the proposed ShuffleMix, $r$ is more sensitive than $\alpha$ to classification performance, which confirm that preservation of a part of features is the key factor of our design owing to its nature of soft dropout on improving generalization.

\begin{table*}[t]
\centering
\caption{Comparison evaluation of AP and mAP on the PASCAL VOC 2007 dataset. The sign of ``*" indicates the re-implemented results via running the source code from the original authors. The ResNet101 model is slightly modified with a global max-pooling after the last layer. The best results are marked in bold, and the second best results are marked in lavender background.}
\label{table:voc-2007}
\resizebox{0.98\linewidth}{!}{
\setlength{\tabcolsep}{1.mm}{
\begin{tabular}{@{}l|cccccccccccccccccccc|c@{}}
\toprule
Methods    & aero & bike & bird & boat & bottle & bus & car & cat & chair & cow & table & dog & horse & motor & person & plant & sheep & sofa & train & tv & mAP \\ \midrule
% ResNet101 \cite{he2016deep} & 99.5 & 97.7 & 97.8 & 96.4 & 65.7 & 91.8 & 96.1 & 97.6 & 74.2 & 80.9 & 85.0 & 98.4 & 96.5 & 95.9 & 98.4 & 70.1 & 88.3 & 80.2 & 98.9 & 89.2 & 89.9  \\
HCP \cite{Wei2016HCPAF} & 98.6 & 97.1 & 98.0 & 95.6 & 75.3 & 94.7 & 95.8 & 97.3 & 73.1 & 90.2 & 80.0 & 97.3 & 96.1 & 94.9 & 96.3 & 78.3 & 94.7 & 76.2 & 97.9 & 91.5 & 90.9  \\
RCP \cite{Wang2016BeyondOP} & 99.3 & 97.6 & 98.0 & 96.4 & 79.3 & 93.8 & 96.6 & 97.1 & 78.0 & 88.7 & 87.1 & 97.1 & 96.3 & 95.4 & \textbf{99.1} & 82.1 & 93.6 & 82.2 & 98.4 & 92.8 & 92.5  \\
RDAR \cite{Wang2017MultilabelIR} & 98.6 & 97.4 & 96.3 & 96.2 & 75.2 & 92.4 & 96.5 & 97.1 & 76.5 & 92.0 & 87.7 & 96.8 & 97.5 & 93.8 & 98.5 & 81.6 & 93.7 & 82.8 & 98.6 & 89.3 & 91.9  \\
RARL \cite{Chen2018RecurrentAR} & 98.6 & 97.1 & 97.1 & 95.5 & 75.6 & 92.8 & 96.8 & 97.3 & 78.3 & 92.2 & 87.6 & 96.9 & 96.5 & 93.6 & 98.5 & 81.6 & 93.1 & 83.2 & 98.5 & 89.3 & 92.0  \\
CoP \cite{Wen2021MultilabelIC} & \textbf{99.9} & 98.4 & 97.8 & \textbf{98.8} & 81.2 & 93.7 & 97.1 & \textbf{98.4} & 82.7 & 94.6 & 87.1 & 98.1 & 97.6 & 96.2 & 98.8 & 83.2 & 96.2 & 84.7 & {\cellcolor{Lavender!100}99.1} & 93.5 & 93.8  \\
ML-GCN \cite{Chen2019MultiLabelIR} & 99.5 & {\cellcolor{Lavender!100}98.5} & {\cellcolor{Lavender!100}98.6} & 98.1 & 80.8 & 94.6 & 97.2 & 98.2 & 82.3 & {\cellcolor{Lavender!100}95.7} & 86.4 & 98.2 & 98.4 & {\cellcolor{Lavender!100}96.7} & {\cellcolor{Lavender!100}99.0} & 84.7 & {\cellcolor{Lavender!100}96.7} & 84.3 & 98.9 & 93.7 & 94.0  \\ \midrule

ResNet101* \cite{he2016deep}  & {\cellcolor{Lavender!100}99.8} & 98.3 & 98.4 & 98.3 & 79.4 & 93.4 & 97.2 & 97.6 & 79.5 & 93.7 & 86.5 & 97.8 & 97.9 & 96.2 & 98.7 & 84.0 & 95.8 & 79.8 & 98.7 & 93.3 & 93.2  \\
ResNet101+ShuffleMix & 99.5 & {\cellcolor{Lavender!100}98.5} & 98.3 & {\cellcolor{Lavender!100}98.5} & {\cellcolor{Lavender!100}81.5} & \textbf{96.0} & {\cellcolor{Lavender!100}97.7} & {\cellcolor{Lavender!100}98.3} & {\cellcolor{Lavender!100}83.3} & \textbf{96.9} & 87.1 & {\cellcolor{Lavender!100}98.3} & {\cellcolor{Lavender!100}98.5} & 96.0 & \textbf{99.1} & \textbf{86.3} & {\cellcolor{Lavender!100}96.7} & \textbf{86.7} & 98.9 & {\cellcolor{Lavender!100}94.8} & {\cellcolor{Lavender!100}94.5}  \\ \midrule

ML-GCN* \cite{Chen2019MultiLabelIR}  & {\cellcolor{Lavender!100}99.8} & 98.3 & \textbf{98.7} & 98.2 & 80.3 & 94.6 & 97.3 & 97.8 & 81.1 & 94.9 & {\cellcolor{Lavender!100}87.2} & 97.9 & 98.0 & 95.8 & 98.8 & 82.7 & 95.8 & 82.1 & 98.6 & 93.1 & 93.5  \\
ML-GCN+ShuffleMix & 99.4 & \textbf{99.0} & \textbf{98.7} & {\cellcolor{Lavender!100}98.5} & \textbf{81.6} & {\cellcolor{Lavender!100}95.6} & \textbf{97.8} & \textbf{98.4} & \textbf{83.6} & \textbf{96.9} & \textbf{88.2} & \textbf{98.5} & \textbf{98.7} & \textbf{96.8} & {\cellcolor{Lavender!100}99.0} & {\cellcolor{Lavender!100}84.9} & \textbf{97.2} & {\cellcolor{Lavender!100}85.2} & \textbf{99.3} & \textbf{95.3} & \textbf{94.6}  \\ \bottomrule
\end{tabular}}
}\vspace{-0.2cm}
\end{table*}

% \usepackage{booktabs}
\begin{table}[t]
\centering
\caption{Effects of different threshold m in the proposed ShuffleMix for multi-label classification on the PASCAL VOC 2007 with ResNet101 and ML-GCN backbone.}
\label{table:voc-m-choice}
\begin{tabular}{@{}l|ccccc@{}}
\toprule
threshold\ (m)\ =                & 0.1   & 0.2   & 0.3   & 0.4   & 0.5   \\ \midrule
ResNet101+ShuffleMix & 94.33 & \textbf{94.54} & 94.38 & 94.45 & 94.13  \\  \midrule
ML-GCN+ShuffleMix    & 94.08 & 94.30  & \textbf{94.62} & 94.63 & 94.32 \\ \bottomrule
\end{tabular}
\end{table}

\begin{figure}[t]
    \centering
    \subfigure{ 
            \includegraphics[width=0.88\linewidth, trim=15 5 55 50,clip]{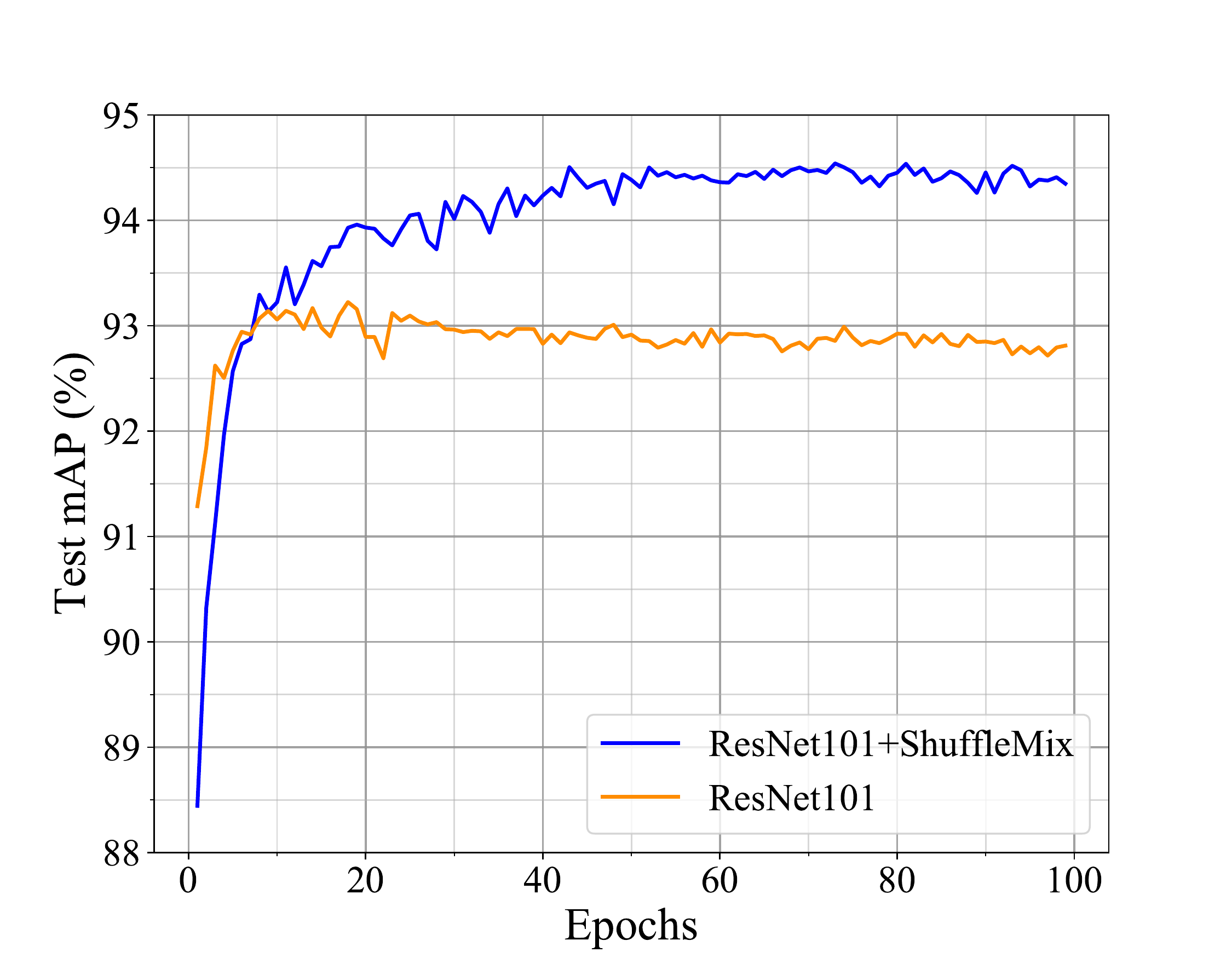}
            \label{fig:VOC_2007}}\\
    %\quad
    \subfigure{ 
            \includegraphics[width=0.88\linewidth, trim=15 5 55 50,clip]{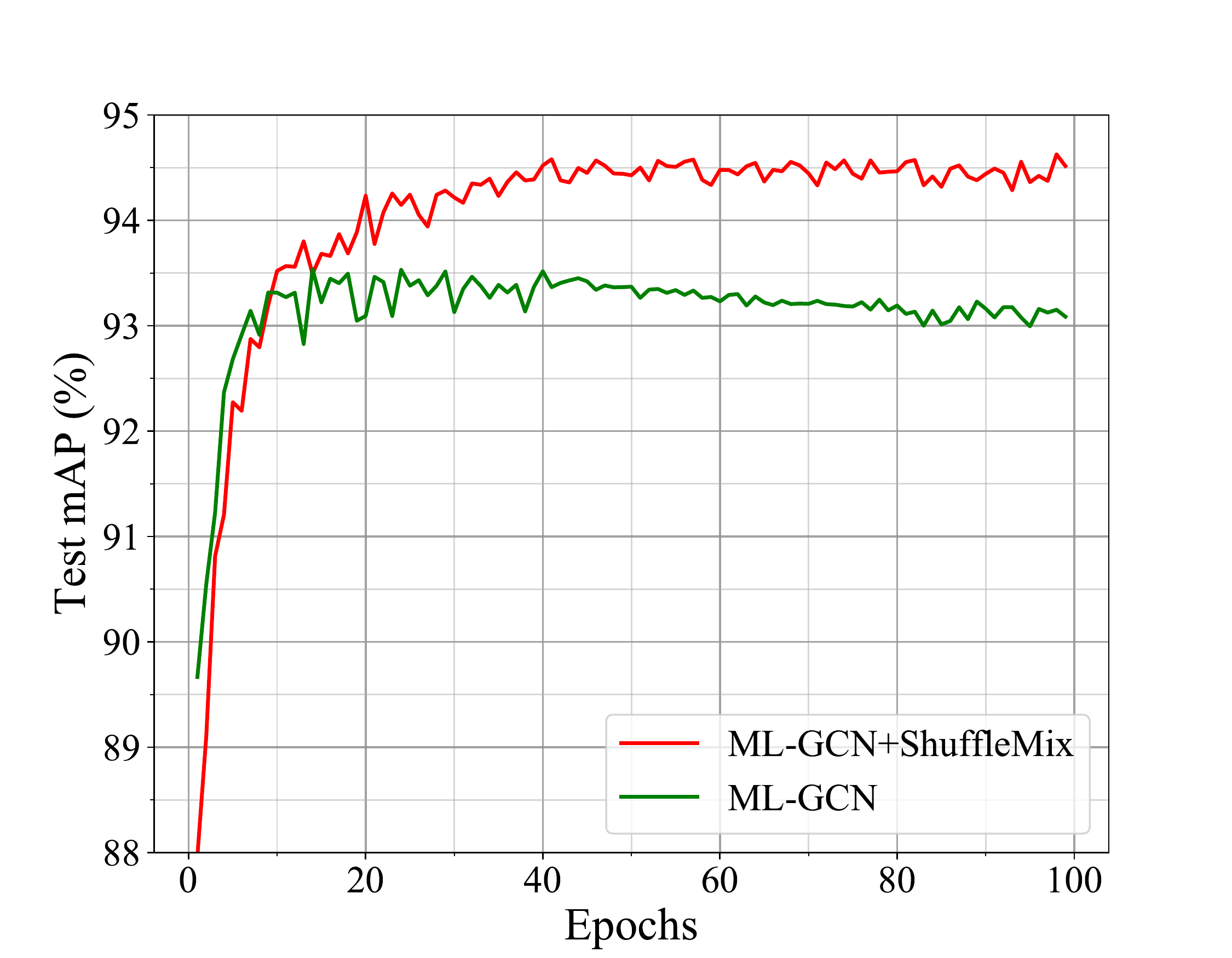}
            \label{fig:COCO}}
    \caption{The test mAP of the ShuffleMix and corresponding networks on the PASCAL VOC 2007 along training epochs. (Best viewed in color)}
    \label{fig:multi_label}%\vspace{-0.2cm}
\end{figure}

\subsection{Evaluation on Multi-Label Image Classification}\label{sec:multi-label}
As the ShuffleMix augments training samples by combining two sampled features during training, it can also be used to improve the representation learning for multi-label classification, as long as the features are similarly trained from the network backbone.
To demonstrate that, we apply the ShuffleMix to the multi-label classification task on the PASCAL VOC 2007 dataset \cite{Everingham2009ThePV}, which contains 5,011 training images and 4,952 testing images from 20 categories. As showed in Fig. \ref{fig:images3}, each image from the PASCAL VOC 2007 usually contains several different categories of objects. 
We follow the same experiment settings in the ML-GCN \cite{Chen2019MultiLabelIR} and implement the ShuffleMix algorithm with ResNet-101 \cite{he2016deep} and ML-GCN \cite{Chen2019MultiLabelIR} networks, respectively.
Motivated by the ML-GCN \cite{Chen2019MultiLabelIR}, the ResNet-101 used here is slightly modified by replacing with a global max-pooling after the last layer, which can help the network get better performance. 
To compare with other related works, we report the results with the average precision (AP) of each category and the mean average precision (mAP) of all categories. 
The hyper-parameters $\alpha=1$, $r=0.5$ and $\mathcal{S}=\{0,1,2,3,4\}$ in the ShuffleMix are set as usual like that stated in Sec. \ref{sec:setting}, and the parameter $m$ in Eq. (\ref{eq:threshold}) will be studied in ablation studies.

The results on the PASCAL VOC 2007 benchmark are reported in Table \ref{table:voc-2007}, where we reproduced the results of ResNet-101 \cite{he2016deep} and ML-GCN \cite{Chen2019MultiLabelIR} under the same experiment settings for a fair comparison. 
It is obvious that the proposed ShuffleMix can improve the performance significantly even with different backbone networks for multi-label classification. 
Concretely, the proposed ShuffleMix with corresponding networks can outperform the ResNet-101 by 1.3\% mAP and the ML-GCN by 1.1\% mAP, respectively. 
It's worth noting that the proposed ShuffleMix with ML-GCN network can obtain 94.6\% mAP, while even with ResNet-101 network, our ShuffleMix can achieve 94.5\% mAP.%, which can verify the effectiveness of the ShuffleMix for multi-label classification.

As illustrated in Fig. \ref{fig:multi_label}, we can find out that the test mAP of our ShuffleMix with baseline networks can be continuously improved with epochs evolve, while the base networks are easy to tend to the state without accuracy improvement.
In our opinion, compared with baseline, the proposed ShuffleMix can continuously provide mixed multi-label data or features for model training. Consequently, the ShuffleMix can effectively improve the training of multi-label classification tasks.% with small training samples.

The hyper-parameter $m$ is designed for labels modification, and aims to control the influence of mixed features and corresponding labels.
Ablation study about $m$ is shown in Table \ref{table:voc-m-choice}, where we can conclude that the best results will be achieved when we set $m$ to a value of 0.2-0.4. 
Our explanation is that too small value of $m$ can cause noise labels into model training, while a large $m$ can filter out labels of important scene objects. %some appropriate labels.

\subsection{Visualization}\label{sec:visualization}

\begin{figure}[htbp]
    \centering
    \includegraphics[width=\linewidth]{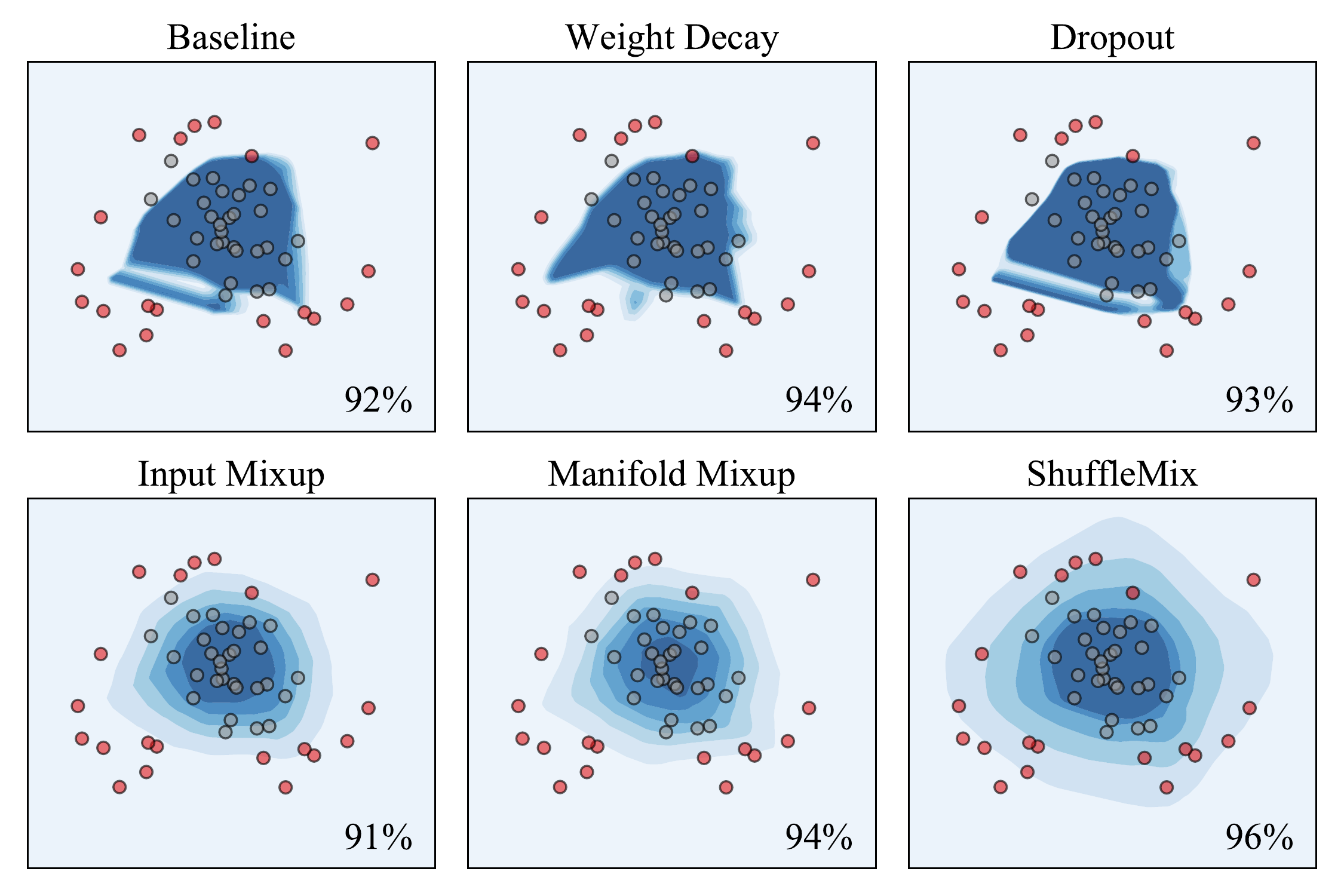}
    \caption{Decision boundaries for different methods on a synthetic dataset of binary classification, which is implemented with the scikit-learn \cite{scikit-learn} package. Classification accuracy of each method is presented in the bottom right of figure. (Best viewed in colors)}
    \label{fig:smooth_boundary}%\vspace{-0.5cm}
\end{figure}

To illustrate advantages of our proposed ShuffleMix method, we further conduct binary classification experiments with several methods, \ie the Weight Decay \cite{Loshchilov2019DecoupledWD}, the Dropout \cite{Srivastava2014DropoutAS}, the Mixup \cite{zhang2018mixup}, and the Manifold Mixup \cite{Verma2019ManifoldMB}, on a synthetic dataset, which is generated with the function of {\em make circles} in scikit-learn \cite{scikit-learn}. 
We use a neural network with four fully-connected (FC) layers and three ReLu modules as our Baseline. The Dropout with dropping rate of 0.2 is just applied to the final FC layer, while the weight decay is set as $5\times10^{-4}$ just for the Weight Decay method. The hyper parameters in Input Mixup, Manifold Mixup, and our ShuffleMix are set the same as stated in Sec. \ref{sec:setting}. 
Meanwhile, we conduct more experiments with the same hyper-parameters on a 3-classes synthetic dataset, %which is similarly generated with the scikit-learn \cite{scikit-learn} package, 
for illustrating main differences among Input Mixup \cite{zhang2018mixup}, Manifold Mixup \cite{Verma2019ManifoldMB}, and our ShuffleMix on representation manifolds (\cf Fig. \ref{fig:boundaries}). 
% The visual results illustrated in Figure \ref{fig:smooth_boundary_3classes} are trained .

Visualization for binary classification is illustrated in Fig. \ref{fig:smooth_boundary}, where we see that mixup style data augmentation methods can always obtain smoother decision boundaries of image classification than typical regularization methods, \eg Weight Decay and Dropout.   
Meanwhile, our ShuffleMix can further obtain the better representation manifolds (ratio of samples falling into the darkest region)
%generalization and smoother decision boundaries 
than the Mixup and the Manifold Mixup, owing to the more flexibility of our ShuffleMix. 
%In our analysis, the randomly selected channel-wise linear operations in our method can gain model generalization via increasing margins to smoother decision boundaries, and are usually more flexible and more robust than the former Manifold Mixup.

%-----------------------------------------------------------------------------------------
\section{Conclusion}\label{sec:conclusion}
In this paper, we propose a novel ShuffleMix feature augmentation method for supervised classification, which is designed in a generalized linear combination of high-level hidden representations.
% rather than linear interpolations in the Manifold Mixup. 
Such a simple yet effective design can improve model generalization owing to its soft dropout-like characteristics and robustness against input noises and data sparsity.
% in view of intrinsic feature perturbation as regularization.
The proposed ShuffleMix is verified its effectiveness not only for single-label image classification but also for multi-label image classification.
%, especially for multi-label image classification.
Moreover, superior performance can further be achieved by combining the ShuffleMix with the noise injection.
%can dynamically augment interpolated feature representation and corresponding label. 
Extensive experiments on several popular datasets with multiple backbone networks can demonstrate consistently superior performance of our method to recent mixup-style data augmentation methods including the state-of-the-art methods.
% In the future, we will further carry out the theoretical analysis of our dynamic feature augmentation for better understanding.
%In the future, we will further carry out the theoretical analysis of such shuffle mix data augmentation for better understanding.

%\section*{Acknowledgements}

%-----------------------------------------------------------------------------------------

% \appendices

% % you can choose not to have a title for an appendix
% % if you want by leaving the argument blank
% \section{}

% % use section* for acknowledgment
% \section*{Acknowledgment}

\ifCLASSOPTIONcaptionsoff
  \newpage
\fi



\bibliographystyle{IEEEtran}
\bibliography{IEEEabrv,tmm22}

\end{document}